\documentclass{article}
\usepackage{float}

 \PassOptionsToPackage{numbers, compress}{natbib}
\usepackage[main, final]{neurips_2026}

\usepackage[utf8]{inputenc} %
\usepackage[T1]{fontenc}    %
\usepackage{hyperref}       %
\usepackage{url}            %
\usepackage{booktabs}       %
\usepackage{amsfonts}       %
\usepackage{nicefrac}       %
\usepackage{microtype}      %
\usepackage{xcolor}         %
\usepackage{subcaption}
\usepackage{wrapfig}
 \usepackage{orcidlink}
 \usepackage{algorithm}
 \usepackage{algpseudocode}
 \usepackage{amsmath, amssymb, amsthm}
\usepackage{adjustbox} %
\usepackage{empheq}
\usepackage[most]{tcolorbox}
\usepackage{multirow}
\usepackage{tcolorbox}

\theoremstyle{plain}
\newtheorem{theorem}{Theorem}[section]

\newtheorem{lemma}[theorem]{Lemma}
\newtheorem{corollary}[theorem]{Corollary}
\theoremstyle{definition}

\newtheorem{assumption}[theorem]{Assumption}
\theoremstyle{remark}

\newtcbox{\eqbox}{
  colback=blue!15,
  colframe=black,
  arc=3mm,
  boxrule=0.8pt,
  left=6pt,
  right=6pt,
  top=6pt,
  bottom=6pt
}

\DeclareMathOperator{\Var}{\mathbb{V}ar}
\DeclareMathOperator{\Cov}{Cov}

\title{Don't Match the Noise: Distribution Matching under Unknown Measurement Error with an Audit Sample}

\author{%
  Arya~Farahi \orcidlink{0000-0003-0777-4618} \quad Ritwik~Vashistha \orcidlink{0009-0004-5711-3222}  \\
  Department of Statistics and Data Science, University of Texas at Austin, Austin, Texas 78712, USA\\
  \texttt{\{arya.farahi, ritwik.v\}@utexas.edu} \\
}

\begin{document}

\maketitle

\vspace{-7mm}
\begin{abstract}
\vspace{-3mm}
Real-world machine learning systems often operate on imperfect measurements: sensors are noisy, annotations are produced by non-experts, and scientific observations are often indirect. In these settings, classic distribution comparison tools such as maximum mean discrepancy (MMD) can be systematically biased, matching artifacts of the measurement process rather than the underlying mechanisms of interest. We study a practical regime in which noisy observations are abundant, but only a small audited subset contains ground-truth measurements. We propose measurement-error-corrected MMD (MEC-MMD), an audit-based estimator that debiases kernel evaluations without requiring a noise model. MEC-MMD is exactly unbiased for the oracle MMD under simple random auditing, admits an explicit variance decomposition, and is consistent for the population MMD, achieving the standard $O_p(N^{-1/2})$ rate when the audit size scales with the dataset. We demonstrate the utility of distributional matching with MEC-MMD loss for both parameter estimation and unsupervised applications, showing that MEC-MMD consistently outperforms MMD and alternative methods. These results provide a principled foundation for robust distribution comparison, simulation-based inference, and learning in noisy real-world machine learning pipelines.\vspace{-5mm}
\end{abstract}

\section{Introduction} \vspace{-2mm}

In many scientific applications, the quantities of interest are not observed directly but only through noisy or imperfect measurements of latent variables. This setting is pervasive across domains including astronomy \citep{farahi2021pope,marshall2021concordance,shy2022incorporating}, remote sensing \citep{congalton2001accuracy}, nutrition sciences \citep{thiebaut2007necessary}, experimental biology \citep{leek2010tackling,mccoy2026tree}, economics \citep{bound1991extent,thurow2025characterizing}, sports analytics \citep{glazer2025beyond}, and differential privacy \citep{jiang2024analysis}. The data-generation process often involves instrument noise, preprocessing artifacts, discretization (e.g., rounding or binning), or the use of proxies. Hence, the analyst observes a proxy  {\footnotesize $\widetilde{Y}$} rather than the scientific target $Y$. Na\"ive procedures that ignore this discrepancy typically learn the distribution of {\footnotesize $\widetilde{Y}$} instead of that of $Y$, leading to biased or scientifically misleading conclusions \citep{bartlett2023marginalised}. This issue is particularly acute in simulation-based inference, where parameters are estimated by matching simulated and observed data distributions \citep{lintusaari2017fundamentals,cranmer2020frontier}. When observations are corrupted, discrepancies computed directly on {\footnotesize $\widetilde{Y}$} favor simulators that reproduce the measurement process rather than the underlying latent mechanism. Properly accounting for measurement error is therefore essential for valid scientific inference. \vspace{-1mm}

In this work, we focus on \emph{measurement error} \citep{carroll2006measurement,buonaccorsi2010measurement}, that differ from data contamination \cite{huber2011robust, ruppert1987robust}. Under contamination \cite[e.g.,][]{farahi2026two, gligic2026robust}, a subset of observations arises from an extraneous distribution; under measurement error, each observation corresponds to the correct latent unit but is recorded imperfectly. Classical approaches to measurement error include deconvolution density estimation \citep{bovy2011extreme,sarkar2014bayesian,melchior2018filling,soloff2025multivariate}, errors-in-variables regression \citep{carroll2006measurement,fuller2009measurement}, simulation-extrapolation methods \citep[SIMEX;][]{cook1994simulation,carroll1999nonparametric,staudenmayer2004local}, and likelihood-based latent-variable models \citep{kelly2007some,sarkar2014bayesian_regression,sereno2016bayesian}. While these methods have led to major theoretical and practical advances, they also face well-known challenges. Deconvolution is often ill-posed and sensitive to assumptions on the noise distribution; errors-in-variables methods typically rely on parametric structure, instrumental variables, or replicate measurements for identifiability; SIMEX can be sensitive to extrapolation error and model misspecification; and likelihood-based approaches require a correctly specified and tractable joint model of the latent variables and noise mechanism, which is difficult to justify in modern high-dimensional or weakly supervised settings. \vspace{-1mm}

Many modern data pipelines provide an additional source of information that is not fully leveraged by these classical frameworks: a small \emph{audited} or \emph{validation} subset in which the latent quantity is observed exactly \citep[e.g.,][]{cartwright2019crowdsourcing,hedderich2021analysing,sakai2023ordering,zrnic2024cross,hsu2025leveraging,shen2025integrating}. One often has access to a large sample of noisy observations {\footnotesize $\widetilde{Y}$} together with a much smaller subset for which both {\footnotesize $\widetilde{Y}$} and the corresponding clean value $Y$ are available, allowing the realized measurement error {\footnotesize $\varepsilon\!=\!\widetilde{Y}\!-\!Y$} to be observed on audited points. This regime (abundant noisy data paired with sparse gold-standard measurements) arises naturally in scientific workflows, for example, through manual annotation, high-precision subsampling, or costly experimental validation (see Table~\ref{tab:noisy_gold_patterns} in Supplementary Materials). It lies between fully supervised settings (where clean labels are widely available) and unsupervised settings (where only proxies are observed), and calls for methods that can combine statistical efficiency from large noisy samples with principled bias correction from limited clean data \citep{congalton2001accuracy,amorim2021two,boe2023issues,angelopoulos2023prediction}. \vspace{-1mm}

We propose to address this problem using the maximum mean discrepancy \citep[MMD;][]{gretton2012kernel}, a kernel-based distance between distributions that is well-suited to likelihood-free testing and inference \cite{gretton2006kernel,park2016k2, mitrovic2016dr, briol2019statistical,cherief2020mmd,bharti2023optimally,gerber2023kernel,pacchiardi2024generalized,vashistha2026convolutional}. The proposed method's key advantage is that it is defined through expectations of kernel evaluations, which enables direct correction for measurement error at the level of the empirical loss. This avoids explicit deconvolution or full likelihood specification while remaining compatible with flexible simulators and gradient-based optimization. Building on this idea, we develop corrected MMD objectives for parameter estimation under unknown measurement error. Assuming access to many noisy observations and a smaller, representative audited subset, we construct debiased empirical MMD estimators that target the latent distribution of $Y$ rather than the proxy distribution of {\footnotesize $\widetilde{Y}$}. The approach is distribution-free, scalable, and readily applicable to modern scientific problems in which the measurements are made with noise. See positioning statement in Supplementary Materials. \vspace{-2mm}
\vspace{-2mm}

\section{Additive Measurement Error with an Audit Sample}
\label{sec:type_ii_problem} \vspace{-2mm}

\paragraph{Background.} Let $\mathcal X \subseteq \mathbb R^d$ be a measurable space on which the probability distributions are defined. We denote $X$, a random variable taking values in $\mathcal{X}$ according to a probability distribution $p$, and by $x \in X$, a realization of $X$. Similarly, we denote $Y$, a random variable taking values in $\mathcal{X}$ according to a probability distribution $q$, and by $y \in Y$, a realization of $Y$.  Let $\mathcal{H}$ be a Reproducing Kernel Hilbert Space (RKHS, \cite{berlinet2011reproducing, muandet2017kernel}) of real-valued functions defined on $\mathcal{X}$. The MMD is then defined as 
\begin{equation}
{\rm MMD}(p,q) = \underset{f \in \mathcal{H},\|f\|_{\mathcal{H}} \leq 1}{\sup}|\mathbb{E}_p(f(X)) - \mathbb{E}_q(f(Y))|. 
\end{equation}
MMD quantifies dissimilarity between two distributions by measuring the distance between their respective mean embeddings in the RKHS. The MMD statistic can be simplified using kernels, which provide a rich structure for analysis and computation. 
\begin{assumption}\label{ass:bounded_kernel}
There is $K<\infty$ with $\sup_{x,x'\in\mathcal X}|k(x,x')|\le K$.
\end{assumption}
Let $k:\mathcal X\times\mathcal X \to \mathbb R$ be a bounded positive definite kernel associated with $\mathcal{H}$ (Assumption~\ref{ass:bounded_kernel}). A closed computational form for ${\rm MMD^2}$ is 
    ${\rm MMD}^2(p,q) = \mathbb{E}_{{X}, {X}'}\left[k\left({X}, {X}'\right)\right]+\mathbb{E}_{{Y}, {Y}'}\left[k\left({Y}, {Y}'\right)\right] -2 \mathbb{E}_{{X}, {Y}}\left[k\left({X}, {Y}\right)\right]$,
with $X,X'\!\overset{iid}{\sim}\!p$ and $Y,Y'\!\overset{iid}{\sim}\!q$. For characteristic kernels~\citep{sriperumbudur2011universality}, it has been shown that ${\rm MMD}(p,q)\!=\!0$ if and only if $p\!=\!q$, thus giving a consistent test \citep{gretton2012kernel}. This property justifies ${\rm MMD}$ as a loss function in learning and inference. See Table~\ref{tab:notation} in the Supplementary Materials for a notation summary.

{\bf Problem setup.}
We study a two-sample problem in which the reference sample is observed without error, while the target sample is observed only through noisy surrogates. Let $X_1,\dots,X_n \overset{\mathrm{iid}}{\sim} p$ and    $Y_1,\dots,Y_N \overset{\mathrm{iid}}{\sim} q$, where $p$ and $q$ are probability distributions on $\mathcal X$. The sample $\{X_i\}_{i=1}^n$ is observed directly. By contrast, the latent target variables $\{Y_j\}_{j=1}^N$ are not fully observed. Instead, for each $j$, the analyst observes a noisy proxy 
\begin{equation}
\label{eq:additive_noise_model}
\widetilde Y_j = Y_j + \varepsilon_j,
\qquad
\varepsilon_j \sim P(\cdot \mid \phi_j),
\qquad
\phi_j \sim \Pi,
\end{equation}
where $\phi_j$ is a unit-specific noise parameter (e.g., a noise scale). Thus, the law of the measurement error may vary across observations, allowing for heteroscedastic or otherwise structured noise, for instance when it depends on unit-level covariates. Such heterogeneity is common in practice \citep[e.g.,][]{chinchilli1996weighted,kupek2002bias,spiegelman2011regression,lewbel2012using,farahi2018xxl,farahi2019mass,waterworth2022obese}. We assume that the tuples $(Y_j,\phi_j,\varepsilon_j)$ are independent across $j$, and independent of the clean sample $\{X_i\}_{i=1}^n$. Our method does not model the noise or any covariate driving it.

{\bf Target Quantity.} Our goal is to estimate or test the squared maximum mean discrepancy between the clean reference distribution $p$ and the latent target distribution $q$,
\begin{equation}
\label{eq:me_mmd_target}
\mathrm{MMD}^2(p,q;k) = \mathbb E[k(X,X')] + \mathbb E[k(Y,Y')] - 2\,\mathbb E[k(X,Y)].
\end{equation}
We recall the fact that when the observed target sample is corrupted by measurement error
\begin{equation}
\mathbb E[k(X,X')] + \mathbb E[k(Y,Y')] - 2\,\mathbb E[k(X,Y)] \neq \mathbb E[k(X,X')] + \mathbb E[k(\widetilde Y,\widetilde Y')] - 2\,\mathbb E[k(X,\widetilde Y)].
\end{equation}
The main challenge is structural. We formalize an estimand that is defined in terms of the latent distribution $q$. MMD depends nonlinearly on pairwise kernel evaluations involving the unobserved latent variables $Y_j$. In contrast to settings with erroneous labels in 0-1 loss \citep{natarajan2013learning}, we pointwise reweight the observed proxies to recover the target quantity and correct the kernel terms directly. 

{\bf Oracle Estimator.} The natural (best case scenario) benchmark would be the usual empirical MMD estimator formed from the clean sample $\{X_i\}_{i=1}^n$ and the latent sample $\{Y_j\}_{j=1}^N$. Define \vspace{-2mm}
\begin{align}
A_{pp}^\star := \sum_{i\neq i'} k(X_i,X_{i'}), \quad
A_{qq}^\star := \sum_{j\neq j'} k(Y_j,Y_{j'}),  \quad
A_{pq}^\star := \sum_{i=1}^n\sum_{j=1}^N k(X_i,Y_j).
\end{align}
The resulting oracle estimator is \vspace{-2mm}
\begin{equation}
\label{eq:me_oracle_mmd}
\widehat{\mathrm{MMD}}_\star^2 = \frac{A_{pp}^\star}{n(n-1)} + \frac{A_{qq}^\star}{N(N-1)} - 2\,\frac{A_{pq}^\star}{nN}.
\end{equation}
This statistic is not directly available because most latent values $Y_j$ are unobserved. Nevertheless, it is the finite-sample target of the correction procedure below. A useful feature of the present setting is that every $\widetilde Y_j$ still corresponds to exactly one genuine latent draw $Y_j\sim q$, even when the measurement error depends on covariates. Thus, the target sample size remains $N$. The difficulty lies entirely in recovering the latent kernel evaluations that appear in $A_{qq}^\star$ and $A_{pq}^\star$. \vspace{-2mm}

\paragraph{Audit Sample.} We further assume access to an \emph{audit sample}, a small subset of target indices for which the latent values are measured exactly. Formally, let $L \subset \{1,\dots,N\}$, where
$|L| = m \ll N$, and suppose that for every $j\in L$ the analyst observes $Y_j$ in addition to $\widetilde Y_j$. Equivalently, the realized measurement error $\varepsilon_j = \widetilde Y_j - Y_j$ is observed on the audited units.  We, hence, assume the analyst has access to $\{\widetilde Y_j\}_{j=1}^N$,
$\{Y_j : j\in L\}$ and $\{X_i\}_{i=1}^n$. 
\begin{assumption}\label{ass:srs_audit}
The audit set  {\small $L\!\subset\!\{1,\dots,N\}$} is a random sample without replacement of fixed size $m$ with $2\le m\le N$, drawn independently of $\{X_i\}_{i=1}^n$ and {\small$\{(Y_j,\widetilde Y_j)\}_{j=1}^N$} where selection is independent of the measurement error. Hence {\small$\Pr(j\in L)=\frac mN$} and  {\small$\Pr(j,j'\in L)\!=\!\frac{m(m-1)}{N(N-1)}$} for {\small$j\!\neq\!j'$}.
\end{assumption} 

\vspace{-2mm}
Unless stated otherwise, expectations of the form $\mathbb E_L[\cdot]$ are taken with respect to this audit-design randomness only, conditional on the realized finite population. The role of the audit sample is to reveal enough latent information to reconstruct, in a design-unbiased manner, the kernel contributions that would have been available if the full target sample had been observed without error. Section~\ref{sec:nonuniform_main} relaxes this to known, unequal inclusion probabilities.
\begin{assumption}\label{ass:asymptotic_regime}
$n,N,m\to\infty$ with $m/N\to\rho\in(0,1]$ and $n/N\to\gamma\in(0,\infty)$.
\end{assumption}

\vspace{-2mm}
\section{Method and Theoretical Guarantees}
\label{sec:theory}
\vspace{-3mm}

While the noisy observations $\widetilde Y_j$ provide a cheap proxy for the latent $Y_j$, $\widehat{\mathrm{MMD}}^2(X,Y,k) \neq \widehat{\mathrm{MMD}}^2(X,\widetilde Y,k)$ even when $p=q$. To construct an estimator of $\mathrm{MMD}^2(p,q,k)$, we begin with the plug-in MMD on $\widetilde Y$, and then use the audited subset to correct it. \vspace{-2mm}

\subsection{Estimating Oracle MMD}

We have the following algebraic identities. For every $j\neq j'$,
\begin{equation}
\label{eq:qq_residual_identity}
k(Y_j,Y_{j'}) = k(\widetilde Y_j,\widetilde Y_{j'})+ \Gamma_{jj'},\qquad j,j'=1,\dots,N,
\end{equation}
where $\Gamma_{jj'} := k(Y_j,Y_{j'}) - k(\widetilde Y_j,\widetilde Y_{j'})$.
Likewise, for every $i,j$,
\begin{equation}
\label{eq:pq_residual_identity}
k(X_i,Y_j) = k(X_i,\widetilde Y_j) + \Lambda_{ij}, \qquad j=1,\dots,N,\ \mathrm{and}\quad i=1,\dots,n,
\end{equation}
where $\Lambda_{ij} := k(X_i,Y_j) - k(X_i,\widetilde Y_j)$.
The quantities $\Gamma_{jj'}$ and $\Lambda_{ij}$ exist for every $j$ and $j'$ but are only observed when $j,j'\in L$. Since these are exact algebraic identities, no approximation is involved. Now, define the corrected $qq$ kernel sum by
\begin{equation}
\label{eq:Aqq_me_hat}
\widehat A_{qq} := \sum_{j\neq j'} k(\widetilde Y_j,\widetilde Y_{j'}) + \frac{N(N-1)}{m(m-1)} \sum_{\substack{j,j'\in L\\ j\neq j'}} \Gamma_{jj'},
\end{equation}
and the corrected $pq$ kernel sum by
\begin{equation}
\label{eq:Apq_me_hat}
\widehat A_{pq} := \sum_{i=1}^n\sum_{j=1}^N k(X_i,\widetilde Y_j) + \frac{N}{m} \sum_{i=1}^n\sum_{j\in L} \Lambda_{ij}.
\end{equation}

Because the denominators are known, the measurement-error-corrected MMD estimator is simply
\begin{tcolorbox}[
colframe=teal!90,
colback=teal!5]
\vspace{-2mm}
\begin{equation}
\label{eq:mmd_me}
\widehat{\mathrm{MMD}}^2_{\mathrm{MEC}} = \frac{A_{pp}^\star}{n(n-1)} + \frac{\widehat A_{qq}}{N(N-1)} - 2\,\frac{\widehat A_{pq}}{nN}
\end{equation}
\vspace{-3mm}
\end{tcolorbox}
\vspace{-2mm}
Computing $\widehat{\mathrm{MMD}}^2_{\mathrm{MEC}}$ takes $O(n^2+N^2+nN)$ kernel evaluations, as for the plug-in estimator, plus $O(m^2+nm)$ for the audit correction. Appendix~\ref{app:details_experiments} reports runtime and memory.
\subsection{Design-Based Guarantees and Asymptotics}

Recall the Horvitz--Thompson identities (Lemma~\ref{lem:me_ht_identities}) under Assumption~\ref{ass:srs_audit} \citep{horvitz1952generalization}.

\begin{theorem}
\label{thm:me_unbiased_sums}
Conditional on the realized samples $\{X_i\}_{i=1}^n$ and
$\{(Y_j,\widetilde Y_j)\}_{j=1}^N$. Under Assumption~\ref{ass:srs_audit}, $\mathbb E_L[\widehat A_{qq}] = A_{qq}^\star$ and 
$\mathbb E_L[\widehat A_{pq}] = A_{pq}^\star$. Consequently,
$\mathbb E_L\!\left[\widehat{\mathrm{MMD}}^2_{\mathrm{MEC}}\right] = \widehat{\mathrm{MMD}}_\star^2$.
\end{theorem}
Theorem~\ref{thm:me_unbiased_sums} proves that $\widehat{\mathrm{MMD}}^2_{\mathrm{MEC}}$ is indeed an unbiased estimator of the oracle $\widehat{\mathrm{MMD}}_\star^2$. Now, we compute the variance. 

\begin{theorem} \label{thm:me_variance_compact}
Conditional on the realized samples $\{X_i\}_{i=1}^n$ and
$\{(Y_j,\widetilde Y_j)\}_{j=1}^N$. Suppose Assumptions ~\ref{ass:srs_audit} and ~\ref{ass:bounded_kernel} hold and $N\ge 4$. Define
\[
\Lambda_j:=\frac1n\sum_{i=1}^n \Lambda_{ij},
\
\bar\Lambda:=\frac1N\sum_{j=1}^N \Lambda_j,
\
\bar\Gamma := \binom{N}{2}^{-1}\sum_{1\le j<j'\le N}\Gamma_{jj'},
\
\bar\Gamma_j := \frac1{N-1}\sum_{j'\neq j}\Gamma_{j\wedge j',\,j\vee j'}.
\]
Let
\[
S_{\mathrm{proj}}^2 := \frac1{N-1}\sum_{j=1}^N \left[ \frac{N-1}{N-2}(\bar\Gamma_j-\bar\Gamma)-(\Lambda_j-\bar\Lambda) \right]^2
\]
and
\[
R_{\mathrm{deg}}^2 := \binom{N}{2}^{-1} \sum_{1\le j<j'\le N} \left[ \Gamma_{jj'}-\bar\Gamma -\frac{N-1}{N-2}(\bar\Gamma_j-\bar\Gamma) -\frac{N-1}{N-2}(\bar\Gamma_{j'}-\bar\Gamma) \right]^2.
\]
Then
\begin{equation}
\label{eq:me_variance_compact}
\Var_L\!\left(\widehat{\mathrm{MMD}}_{\mathrm{MEC}}^2\right) = \frac{4(N-m)}{mN}\,S_{\mathrm{proj}}^2 + \frac{2(N-m)(N-m-1)}{m(m-1)(N-2)(N-3)}\,R_{\mathrm{deg}}^2.
\end{equation}
\end{theorem}

The theorem shows that the audit-design variance of the MEC estimator separates cleanly into a first-order projection term and a degenerate second-order remainder, exactly mirroring a Hoeffding decomposition for U-statistics. The dominant contribution is the finite-population variance $S_{\mathrm{proj}}^2$ of the projected unit-level correction $\frac{N-1}{N-2}(\bar\Gamma_j-\bar\Gamma)-(\Lambda_j-\bar\Lambda)$, which captures how much each audited point changes the corrected $qq$ and $pq$ kernel terms. This leading component scales as $O(m^{-1})$, exactly as in standard finite-population mean estimation. The second component, $R_{\mathrm{deg}}^2$, measures residual pairwise interaction. It is not reducible to unit-level effects because its coefficient is order $m^{-2}$ and $R_{\mathrm{deg}}^2$ is bounded under Assumption~\ref{ass:bounded_kernel}, therefore, it is asymptotically negligible.

\begin{corollary}
\label{cor:me_variance_asymptotic}
Under the conditions of Theorem~\ref{thm:me_variance_compact} and $\frac{m}{N}\to \rho\in(0,1]$, we get
\begin{equation}
\label{eq:me_variance_asymptotic}
\Var_L\!\left(\widehat{\mathrm{MMD}}_{\mathrm{MEC}}^2\right) = \frac{4(1-\rho)}{\rho N}\,S_{\mathrm{proj}}^2 + \frac{2(1-\rho)^2}{\rho^2 N^2}\,R_{\mathrm{deg}}^2 + o(N^{-1}).
\end{equation}
In particular, the leading term is typically
\[
\Var_L\!\left(\widehat{\mathrm{MMD}}_{\mathrm{MEC}}^2\right)
\approx \frac{4(N-m)}{mN}\,S_{\mathrm{proj}}^2.
\]
\end{corollary}
Under Assumption~\ref{ass:bounded_kernel}, $S_{\mathrm{proj}}^2\le 134K^2$ and $R_{\mathrm{deg}}^2\le 4K^2$ for all $N\ge 4$ (Corollary~\ref{cor:me_variance_asymptotic}). Conceptually, the results here imply that the measurement-error problem behaves, to first order, like estimating an average unit-level correction, even though the underlying MMD functional is nonlinear in the latent sample. This makes the corrected procedure much more interpretable: to first order, increasing the audit size reduces variance at the familiar $1/m$ rate, while genuinely nonlinear pairwise effects appear only as a smaller second-order refinement.

\subsection{Oracle Recovery and Consistency}
\label{sec:me_oracle_recovery}

\begin{theorem}[Design consistency and CLT]
\label{thm:me_design_clt}
Suppose Assumptions~\ref{ass:srs_audit}, \ref{ass:asymptotic_regime}, and \ref{ass:bounded_kernel} hold. Conditional on the realized finite population
$\{(Y_j,\widetilde Y_j)\}_{j=1}^N$ and $\{X_i\}_{i=1}^n$.
Then
\begin{equation}
\label{eq:me_normalized_rates}
\frac{\widehat A_{pq}-A_{pq}^\star}{nN}
=
O_{\mathbb P_L}(m^{-1/2}),
\qquad
\frac{\widehat A_{qq}-A_{qq}^\star}{N(N-1)}
=
O_{\mathbb P_L}(m^{-1/2}).
\end{equation}
Moreover, if the corresponding finite-population asymptotic variances converge to finite limits, then the vector
\[
\sqrt m
\begin{pmatrix}
(\widehat A_{pq}-A_{pq}^\star)/(nN)\\[3pt]
(\widehat A_{qq}-A_{qq}^\star)/(N(N-1))
\end{pmatrix}
\]
is asymptotically Gaussian under the audit-design randomness. Consequently,
\begin{equation}
\label{eq:me_oracle_clt}
\sqrt m\left(
\widehat{\mathrm{MMD}}^2_{\mathrm{MEC}}
-
\widehat{\mathrm{MMD}}_\star^2
\right)
\Rightarrow
\mathcal N(0,V_{\mathrm{design}})
\end{equation}
for some finite $V_{\mathrm{design}}\ge 0$ given by the asymptotic variance of the corresponding linear combination.
\end{theorem}

Theorem~\ref{thm:me_design_clt} is essentially a finite population CLT for HT estimators \citep{serfling1974probability} and shows that, under simple random auditing, the correction error of MEC is asymptotically negligible at the standard $m^{-1/2}$ rate and satisfies a design-based CLT around the oracle empirical MMD. In particular, once the audit size $m$ grows, the dominant uncertainty induced by partial auditing behaves like ordinary sampling noise from the audited subset rather than a more pathological nonlinear error arising from measurement correction. Practically, it implies that inference for the corrected statistic can be based on the same first-order Gaussian approximations that underlie classical survey-sampling and U-statistic theory, with the oracle empirical MMD serving as the finite-sample target and the audit design contributing a transparent additional variance term. A simulation in Appendix~\ref{app:details_experiments} (Table~\ref{tab:clt_check}) confirms the unbiasedness and the variance formula and shows that the normal approximation is standardized by the true variance.

\begin{theorem}[Consistency]
\label{thm:me_consistency_population}
Suppose Assumptions~\ref{ass:additive_me_model}, \ref{ass:srs_audit},
\ref{ass:asymptotic_regime}, and \ref{ass:bounded_kernel} hold. Then
\begin{equation}
\label{eq:me_population_consistency}
\widehat{\mathrm{MMD}}^2_{\mathrm{MEC}}
\;\xrightarrow[]{\mathbb P}\;
\mathrm{MMD}^2(p,q;k),
\end{equation}
where the probability is with respect to all sources of randomness
$(X,Y,\phi,\varepsilon,L)$.

More precisely,
\begin{equation}
\label{eq:me_population_rate}
\widehat{\mathrm{MMD}}^2_{\mathrm{MEC}} - \mathrm{MMD}^2(p,q;k) = O_{\mathbb P}(n^{-1/2}+N^{-1/2}+m^{-1/2}).
\end{equation}
Under Assumption~\ref{ass:asymptotic_regime}, this simplifies to
\begin{equation}
\label{eq:me_population_rate_N}
\widehat{\mathrm{MMD}}^2_{\mathrm{MEC}} - \mathrm{MMD}^2(p,q;k) = O_{\mathbb P}(N^{-1/2}).
\end{equation}
\end{theorem}

The theorem decomposes the estimation error into an audit term of order $m^{-1/2}$ and the usual empirical MMD sampling terms of order $n^{-1/2}$ and $N^{-1/2}$. Under Assumption~\ref{ass:asymptotic_regime}, all three are of order $N^{-1/2}$. Thus, if the audit size grows proportionally with the target sample size, partial auditing, while increasing variance, asymptotically does not degrade the first-order convergence rate. 
\vspace{-2mm}

\vspace{-2mm}
\section{Experiments} \label{sec:experiments}
\vspace{-2mm}

\begin{figure}
    \centering
    \includegraphics[width=0.32\linewidth]{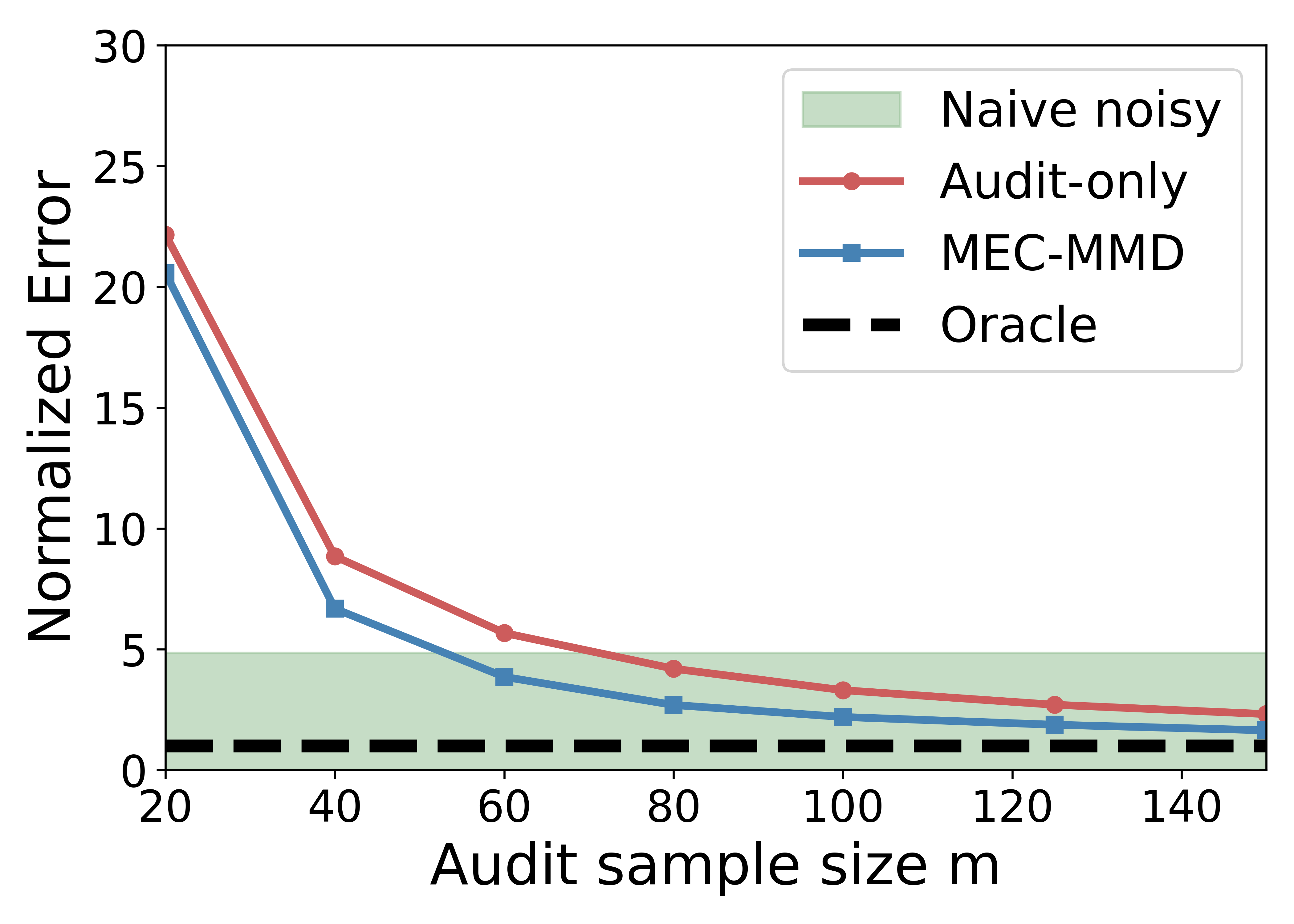}
    \includegraphics[width=0.32\linewidth]{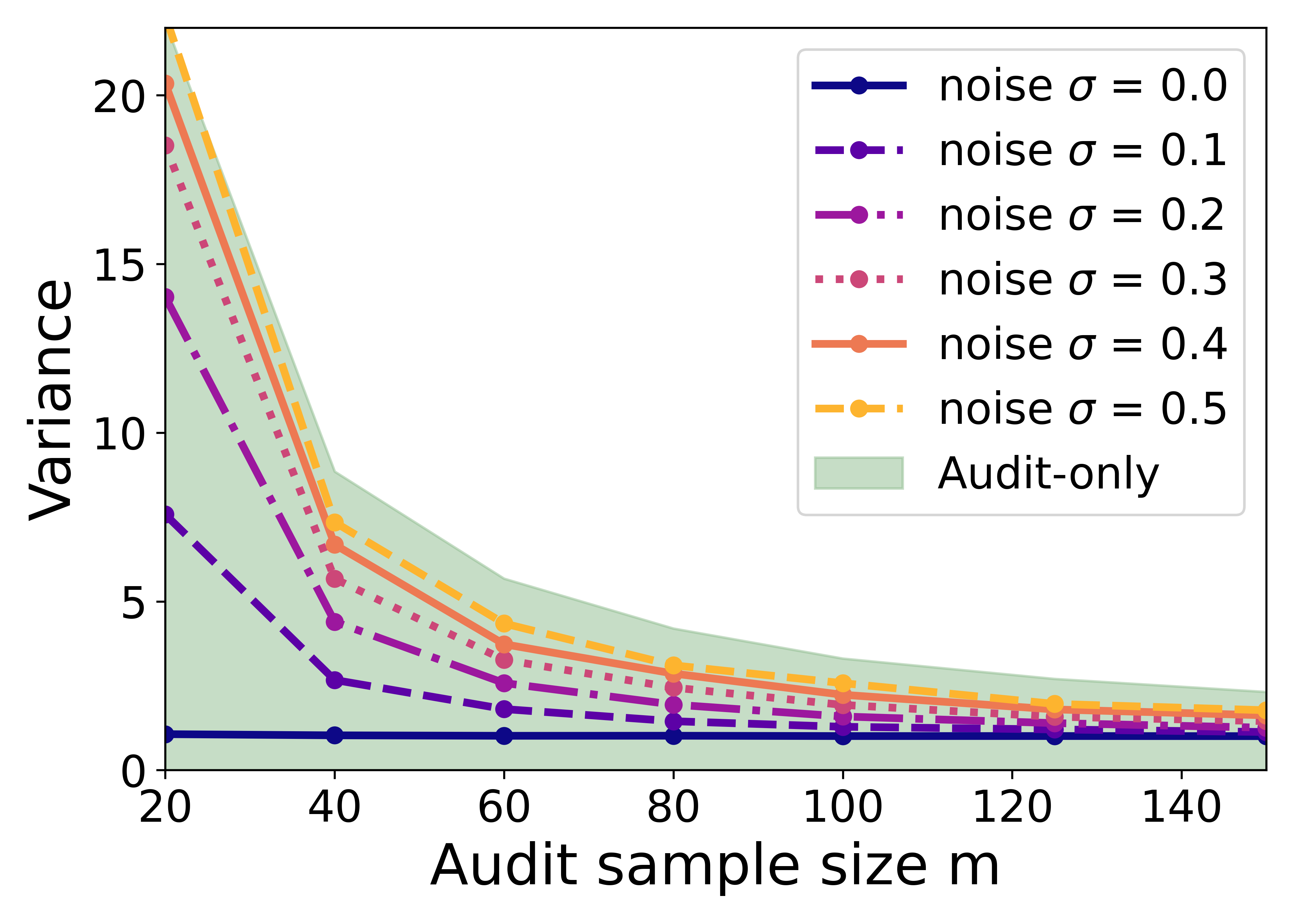}
    \includegraphics[width=0.32\linewidth]{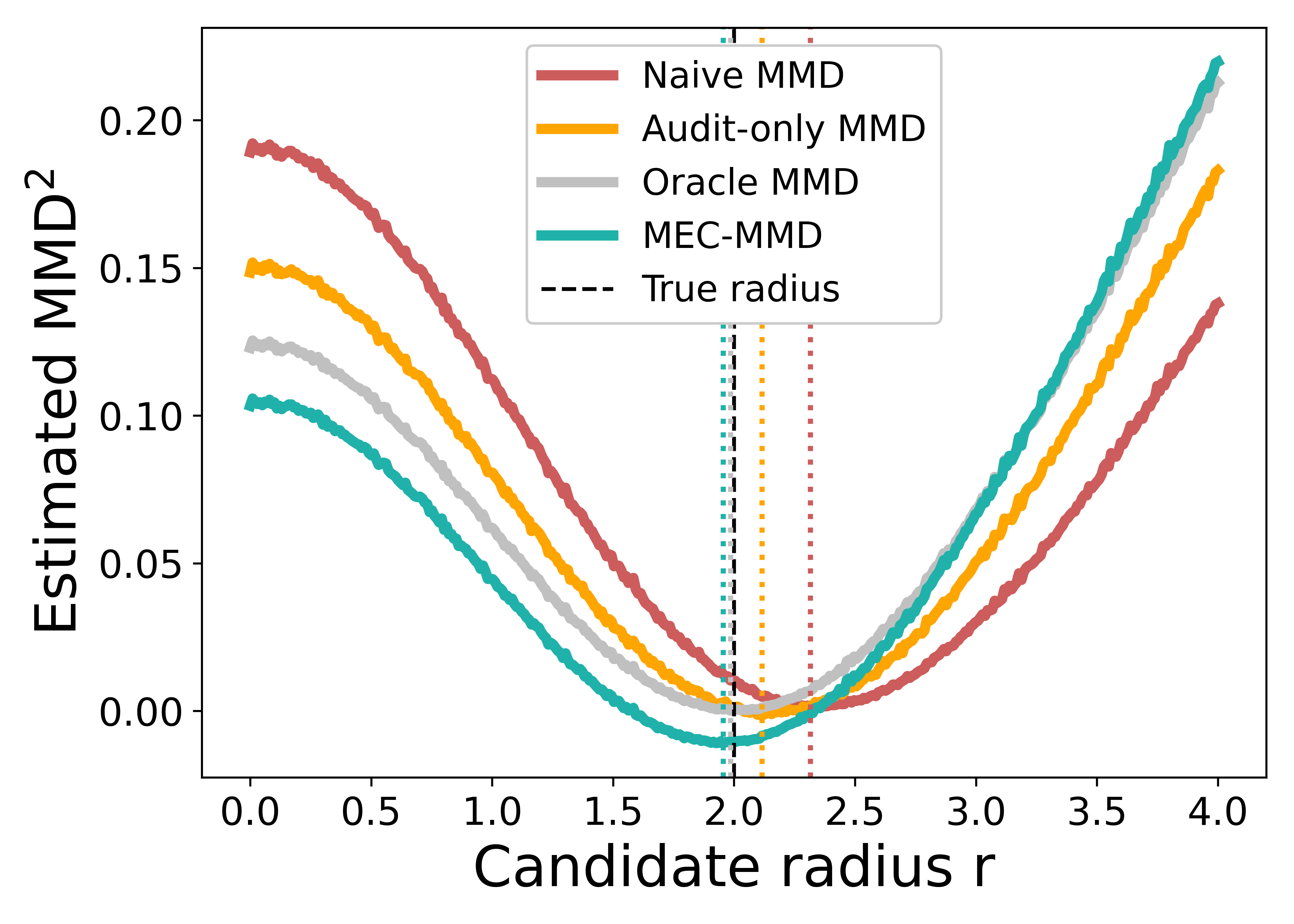}
    \vspace{-1mm}
    \caption{{\bf Left Panel.} Normalized error (MSE divided by the oracle's) versus audit size. The naive estimator is low-variance but biased, the audit-only estimator is unbiased but high-variance, and MEC-MMD removes bias while reducing variance. {\bf Middle Panel.} Effect of measurement noise on MMD estimation variance for several noise levels $\sigma$. The shaded region is the audit-only variance, independent of $\sigma$. {\bf Right Panel.}  MEC-MMD closely matches the oracle objective and correctly identifies the true parameter with lower bias and variance. See Supplementary Materials for details.}
        \vspace{-5mm}
    \label{fig:me_main_result}
\end{figure}

\paragraph{Experiment 1 -- Bias--Variance Trade-off.} We study the bias--variance trade-off of measurement-error correction in a one-dimensional Gaussian example. We generate $X_{1:n}\sim\mathcal N(3,1)$ and $Y_{1:n}\sim\mathcal N(0,1)$ with $n\!=\!500$, but observe only noisy targets $\widetilde Y_j\! =\! Y_j + \varepsilon_j$, where $\varepsilon_j\!\sim\!\mathcal N(0.1,0.4^2)$. Figure~\ref{fig:me_main_result} (left) compares four estimators: the oracle using the full latent sample $Y$, the naive estimator using noisy proxies $\widetilde Y$, the audit-only estimator using only $m$ audited latent points, and the proposed MEC-MMD estimator. The naive estimator has low variance but is biased, whereas the audit-only estimator removes this bias at the cost of substantially higher variance because $m\!\ll\!n$ (Figure  \ref{fig:me_bias_variance_appendix}). MEC-MMD combines the noisy sample with an audit-based correction, removing bias while retaining much of the efficiency of the full sample. See Supplementary Materials for the experiment details. \vspace{-3mm}

\paragraph{Experiment 2 -- Measurement Error Effect.} We study how measurement noise affects the efficiency of MEC-MMD. Figure~\ref{fig:me_main_result} (middle) shows the normalized variance of MEC-MMD versus audit size $m$ across several noise levels. Variance decreases with $m$, consistent with the $m^{-1}$ scaling in Corollary~\ref{cor:me_variance_asymptotic}, but increases with the noise magnitude because larger corrections are required. Increasing measurement noise raises the variance of MEC-MMD, since the correction must compensate for larger discrepancies between latent and noisy kernel evaluations. For moderate noise ($\sigma\le 0.5$) MEC-MMD is more efficient than the audit-only estimator, but this advantage shrinks as $\sigma$ grows. Once the noise is comparable to the signal scale ($\sigma\gtrsim 1$ here), the correction terms carry little information about the latent kernel values and mostly add variance, so the audit-only estimator is preferable. As the noise level approaches zero, MEC-MMD converges to the oracle variance. The results provide evidence that MEC-MMD remains effective under moderate measurement error. \vspace{-3mm}

\paragraph{Experiment 3 -- Parameter Estimation.} We study parameter estimation in a geometric model where the latent target distribution is a noisy circle with unknown radius $r_\star$. Latent points are drawn from random angles and perturbed radii $R\!=\!r_\star\!+\!\eta$, with $\eta\!\sim\!\mathcal N(0,\sigma_r^2)$, but the analyst observes only noisy proxies $\widetilde Y\!=\!Y\!+\!\varepsilon$, except for a small audited subset with exact latent values. To estimate $r_\star$, we generate candidate samples from noisy circles with radius $r$ and evaluate empirical MMD between each candidate model and the observed data. Figure~\ref{fig:me_main_result} (right) compares the resulting objective curves for four estimators: oracle, naive, audit-only, and MEC-MMD. The naive estimator is biased by measurement error and therefore attains its minimum at a shifted radius. The audit-only estimator removes this bias, but it has high variance because it relies on only a small audit sample. MEC-MMD combines the full noisy sample with an audit-based correction, yielding an objective whose minimum closely tracks the oracle while remaining substantially more stable than audit-only estimation. This illustrates MEC-MMD's potential in likelihood-free model parameter estimation settings \cite{briol2019statistical}. \vspace{-3mm}

\begin{figure}
    \centering
    \includegraphics[width=0.98\linewidth]{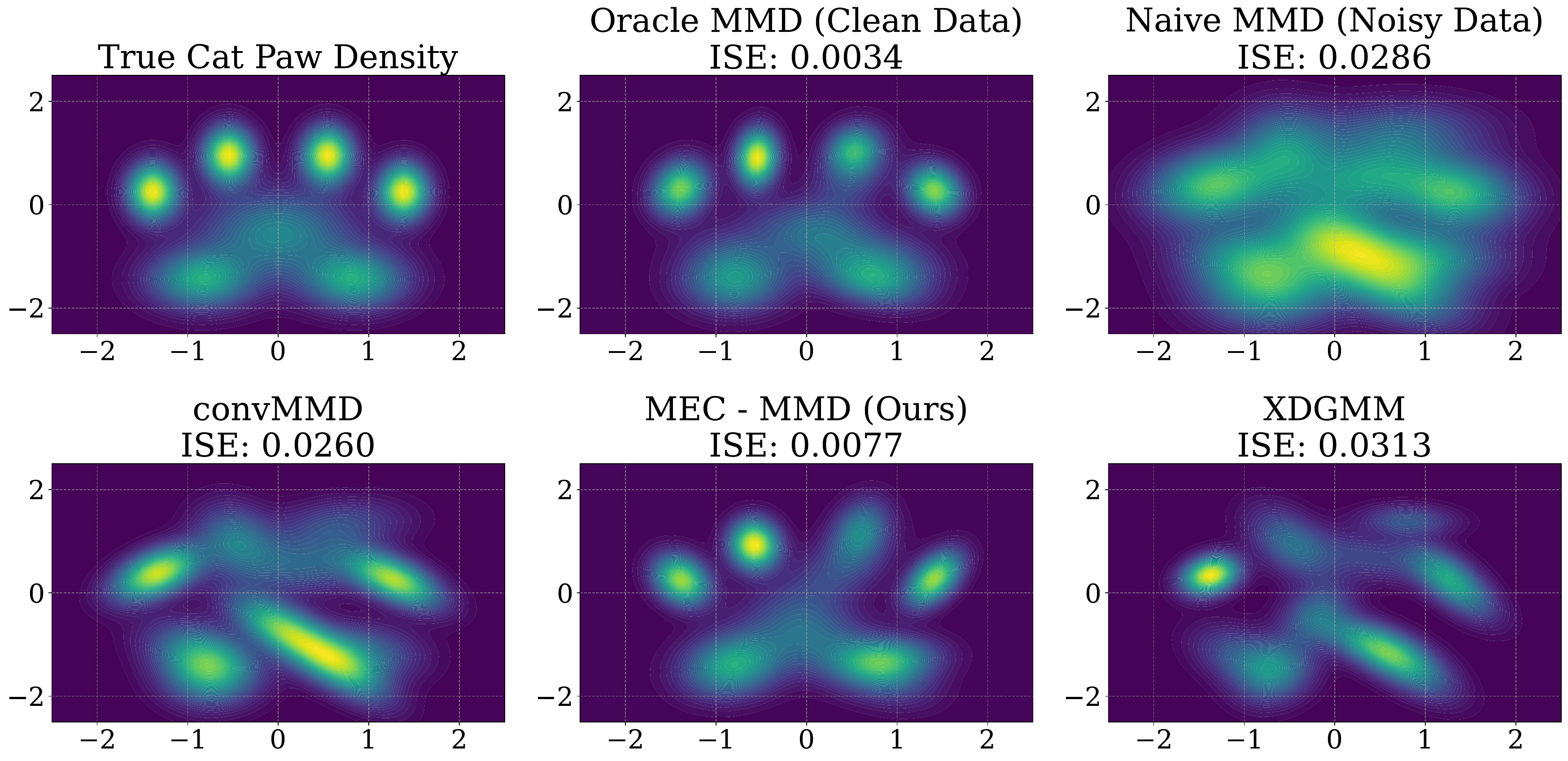}
        \vspace{-3mm}
    \caption{Comparison of Density Estimation in the presence of mean-zero Gaussian noise.}
    \vspace{-5mm}
    \label{fig: paw_density}
\end{figure}

\paragraph{Experiment 4a -- Density Estimation.}

We illustrate the performance of our method for density estimation. The true latent distribution resembles a paw of an animal that was generated using a mixture of Gaussians. We consider three classes of heteroskedastic noise models in our experiment: (1) Gaussian, (2) Laplace, and (3) dependent measurement noise. In the case of Gaussian and Laplace measurement noise, the scale parameters are sampled from a uniform distribution, whereas in the case of dependent measurement noise, the Laplace distribution's scale parameters depend on the absolute values of the variables. We fix the sample size ($N\!=\!2000$) and vary the audit sample size $m$ to compare the performance of our method for estimating the density in terms of Integrated Square Error (ISE). All MMD-based methods fit the same model, a 10-component Gaussian mixture model. The model supplies the clean sample $X$ of Section~\ref{sec:type_ii_problem} (via reparameterised draws), and the methods differ only in the data side of the loss. Oracle uses the clean $Y$, Naive uses $\widetilde Y$, MEC-MMD uses $\widetilde Y$ with the audit correction of Eq.~\eqref{eq:mmd_me}, and convMMD adds resampled audit residuals to the model draws before comparing with $\widetilde Y$. The fitted mixture is the density estimate, and ISE is computed on a grid against the true density (details in Appendix~\ref{app:details_experiments}).
Figure~\ref{fig: paw_density} and Table~\ref{tab: paw} report the experimental results. Figure~\ref{fig: paw_density} shows the estimated density from a single run under Gaussian noise with audit sample size $m=400$, comparing our method against oracle MMD (best-case) \cite{briol2019statistical}, convMMD \cite{vashistha2026convolutional}, XDGMM \cite{bovy2011extreme}, and naive MMD trained directly on noisy data without noise correction. Table~\ref{tab: paw} summarizes performance across 10 runs for varying audit sizes and noise distributions. MEC-MMD improves consistently as the audit sample size increases, progressively closing the gap to oracle performance. Across all three noise regimes, it substantially outperforms both XDGMM and naive MMD. Notably, for larger audit samples ($m \geq 300$), MEC-MMD also surpasses convMMD and achieves the lowest ISE among all non-oracle methods. The gains are especially pronounced under Gaussian and dependent noise, highlighting the method’s ability to recover the underlying density accurately even in complex, observation-dependent heteroscedastic settings.\vspace{-4mm}

\begin{table}[ht]
\centering
\caption{Mean and Standard Error of ISE across varying noise distributions and audit sample sizes.}
\resizebox{\textwidth}{!}{%
\begin{tabular}{|l|l|c|cccc|}
\toprule
Noise & $m$ & Oracle & MEC-MMD (Ours) & Na\"ive MMD \cite{briol2019statistical} & convMMD \cite{vashistha2026convolutional} & XDGMM \cite{bovy2011extreme} \\
\midrule
 \multirow{3}{*}{Gaussian} & 200 & $0.0025 \pm 0.0002$ & $0.0250 \pm 0.0014$ & $0.0292 \pm 0.0004$ & $\mathbf{0.0225 \pm 0.0015}$ & $0.0293 \pm 0.0033$ \\
 & 300 & $0.0025 \pm 0.0002$ & $\mathbf{0.0160 \pm 0.0017}$ & $0.0292 \pm 0.0004$ & $0.0226 \pm 0.0016$ & $0.0288 \pm 0.0027$ \\
 & 400 & $0.0025 \pm 0.0002$ & $\mathbf{0.0117 \pm 0.0011}$ & $0.0292 \pm 0.0004$ & $0.0228 \pm 0.0016$ & $0.0287 \pm 0.0026$ \\
\midrule
 \multirow{3}{*}{Laplace} & 200 & $0.0032 \pm 0.0002$ & $0.0227 \pm 0.0020$ & $0.0238 \pm 0.0006$ & $\mathbf{0.0150 \pm 0.0009}$ & $0.0387 \pm 0.0061$ \\
 & 300 & $0.0032 \pm 0.0002$ & $\mathbf{0.0162 \pm 0.0016}$ & $0.0238 \pm 0.0006$ & $\mathbf{0.0151 \pm 0.0008}$ & $0.0382 \pm 0.0051$ \\
 & 400 & $0.0032 \pm 0.0002$ & $\mathbf{0.0104 \pm 0.0010}$ & $0.0238 \pm 0.0006$ & $0.0150 \pm 0.0008$ & $0.0392 \pm 0.0044$ \\
\midrule
 \multirow{3}{*}{Dependent} & 200 & $0.0033 \pm 0.0003$ & $0.0187 \pm 0.0018$ & $\mathbf{0.0173 \pm 0.0004}$ & $0.0188 \pm 0.0013$ & $0.0710 \pm 0.0060$ \\
 & 300 & $0.0033 \pm 0.0003$ & $\mathbf{0.0125 \pm 0.0009}$ & $0.0173 \pm 0.0004$ & $0.0182 \pm 0.0010$ & $0.0708 \pm 0.0044$ \\
 & 400 & $0.0033 \pm 0.0003$ & $\mathbf{0.0091 \pm 0.0009}$ & $0.0173 \pm 0.0004$ & $0.0182 \pm 0.0011$ & $0.0699 \pm 0.0036$ \\
\bottomrule
\end{tabular}%
}
\vspace{-4mm}
\label{tab: paw}
\end{table}

\paragraph{Experiment 4b -- Non-uniform audits.} We repeat Experiment~4 with an audit drawn with unequal inclusion probabilities. Each mixture component $k$ receives a multiplier $c_k$ (log-uniform on $[0.3,3]$, re-centred to geometric mean one, the same for all seeds); units are audited independently with probability $\pi_j\propto c_{k(j)}$, so the realised $m$ is random with $\mathbb E[m]\approx m_{\rm target}$, and the $\pi_j$ are known. MEC-MMD-IPW uses the H\'ajek weights $1/\pi_j$ (Section~\ref{sec:nonuniform_main}) and convMMD and XDGMM receive the same audited samples without weights. Table~\ref{tab:biased_audit} shows that the weighted correction remains effective under the biased design. At $m=200$ the reweighting adds variance and MEC-MMD-IPW is on par with naive MMD ($0.030$) and slightly behind XDGMM ($0.028$) and convMMD ($0.024$). From $m=300$ it attains the lowest ISE among all non-oracle methods, and its error keeps decreasing with $m$, narrowing the gap to the oracle ($0.004$). \vspace{-3mm}

\begin{table}[h]
\centering
\caption{ISE under a non-uniform (stratified) audit with known inclusion probabilities (Gaussian noise, $N=2000$; mean $\pm$ standard error over 10 seeds). MEC-MMD-IPW uses inverse-probability (H\'ajek) weights; bold marks the best non-oracle method.}
\label{tab:biased_audit}
\resizebox{\textwidth}{!}{%
\begin{tabular}{cccccc}
\toprule
$m$ & Oracle & MEC-MMD-IPW (Ours) & Na\"ive MMD & convMMD & XDGMM \\
\midrule
200 & $0.0038 \pm 0.0014$ & $0.0304 \pm 0.0053$ & $0.0298 \pm 0.0009$ & $\mathbf{0.0241 \pm 0.0019}$ & $0.0282 \pm 0.0034$ \\
300 & $0.0038 \pm 0.0014$ & $\mathbf{0.0203 \pm 0.0026}$ & $0.0298 \pm 0.0009$ & $0.0241 \pm 0.0022$ & $0.0275 \pm 0.0029$ \\
400 & $0.0038 \pm 0.0014$ & $\mathbf{0.0166 \pm 0.0025}$ & $0.0298 \pm 0.0009$ & $0.0238 \pm 0.0022$ & $0.0273 \pm 0.0029$ \\
\bottomrule
\end{tabular}%
}
\end{table}
\paragraph{Experiment 5 -- Denoising with Empirical Bayes in the Presence of Outliers.}

We now demonstrate the application of MEC-MMD in denoising using empirical Bayes in a setting where the noise distribution is unknown to the user, in addition to a small amount of contamination. Concretely, we observe $N=2000$ noisy measurements $X_i=\theta_i+\varepsilon_i$ drawn from some latent distribution $p_\theta$, but the noise levels $\sigma_i$ are not provided. Instead, we assume access to a small audit sample of $m=300$ paired observations $\left\{\left(\theta_i, X_i\right)\right\}_{i=1}^{m}$ drawn from the same data-generating process; the remaining $N-m$ training points are unpaired noisy observations only. Our goal is to learn a density $q_{\widehat{\theta}_N} \approx p_\theta$ and use it as a data-driven prior to estimate the posterior mean $\mathbb{E}\left[\theta_i \mid X_i\right]$ via empirical Bayes. Latent points are corrupted by heteroscedastic Gaussian noise with per-sample standard deviations drawn uniformly, and $3 \%$ of the latent distribution is contaminated with outliers to assess robustness. See Appendix~\ref{app:details_experiments} for details and a sensitivity analysis of the kernel bandwidth.

\begin{figure}[H]
    \centering
    \includegraphics[width=0.90\linewidth]{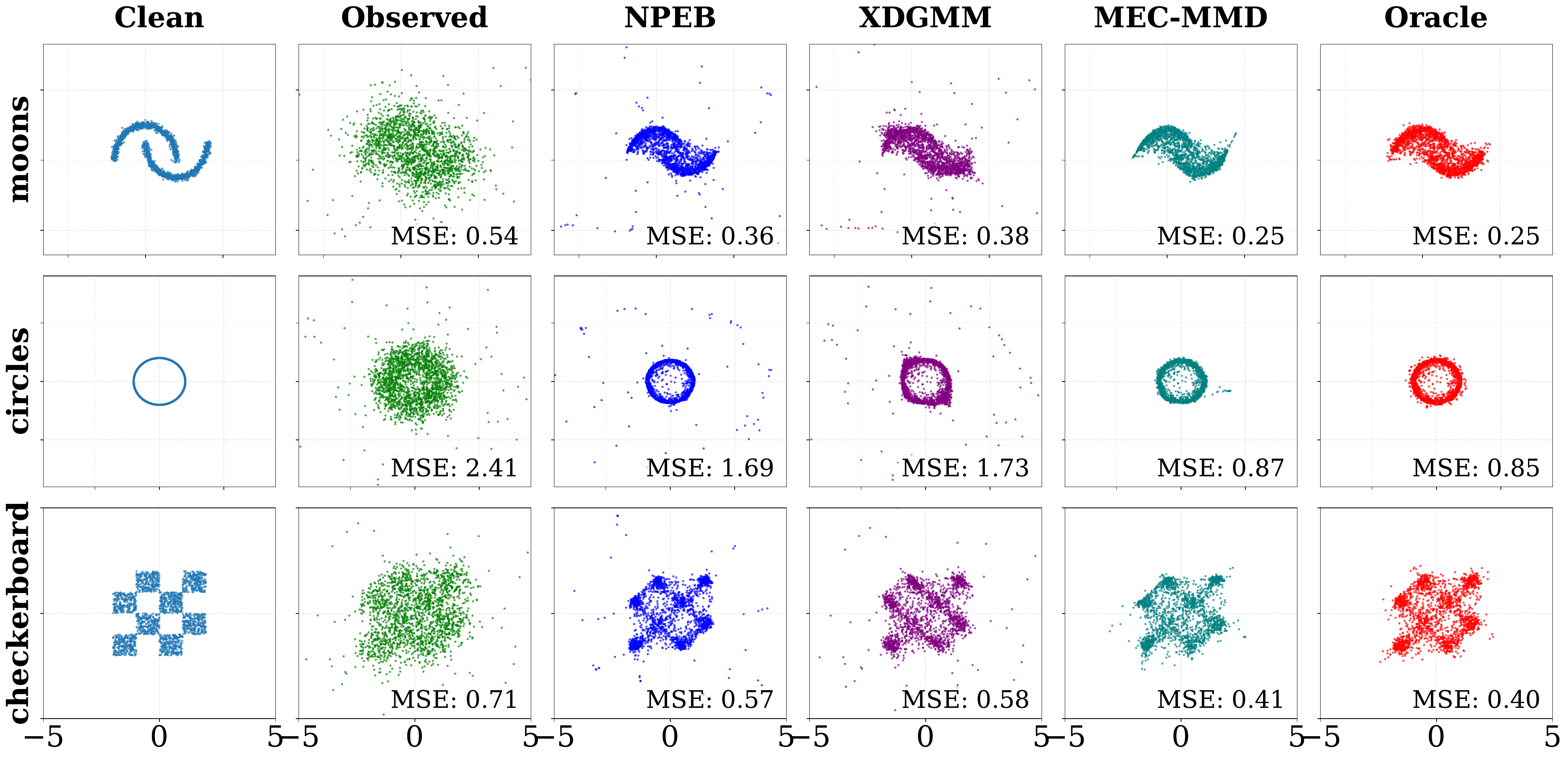}
    \vspace{-1.5mm}
    \caption{A random run of denoising on synthetic data.}
    \label{fig:2d-exp}
    \vspace{-6mm}
\end{figure}

We evaluate on three synthetic datasets (Figure \ref{fig:2d-exp}). We compare against two standard deconvolution baselines: NPEB \cite{soloff2025multivariate}, a nonparametric empirical Bayes method based on the NPMLE, and XDGMM \cite{bovy2011extreme}, which fits a Gaussian-mixture prior. Since both baselines require a noise level as input, we supply them with a single global estimate $\widehat{\sigma}=\operatorname{std}\left(\theta_i-X_i\right)$ computed from the audit pairs; audit observations themselves are treated as exactly observed $\left(\sigma=10^{-6}\right)$. An Oracle baseline is included, which uses the true latent distribution estimated via KDE from a large, clean sample and thus provides an approximate lower bound on denoising error. Results are evaluated on a held-out test set of 2000 points unseen during training, with results averaged across five random seeds. \vspace{-2mm}

\begin{wraptable}{r}{0.55\textwidth}
\vspace{-12pt}
\centering
\small
\centering
\caption{Performance comparison on synthetic datasets.}
\label{tab:detailed_results}
\vspace{-2mm}
\resizebox{0.55\textwidth}{!}{
\begin{tabular}{llcc}
\toprule
\textbf{Dataset} & \textbf{Method} & \textbf{MSE ($\downarrow$)} & \textbf{W1 Distance ($\downarrow$)} \\ \midrule
\multirow{4}{*}{Moons} 
    & NPEB       & $0.3394 \pm 0.0139$ & $0.1187 \pm 0.0102$ \\
    & XDGMM      & $0.3511 \pm 0.0144$ & $0.1600 \pm 0.0094$ \\
    & MEC-MMD  & $\mathbf{0.2546 \pm 0.0041}$ & $\mathbf{0.0957 \pm 0.0086}$ \\ 
    & Oracle     & \textit{0.2524 ± 0.0042} & \textit{0.0398 ± 0.0087} \\
   \midrule
\multirow{4}{*}{Circles} 
    & NPEB       & $1.6596 \pm 0.0805$ & $0.3439 \pm 0.0257$ \\
    & XDGMM      & $1.6609 \pm 0.0668$ & $0.3746 \pm 0.0298$ \\
    & MEC-MMD  & $\mathbf{0.8929 \pm 0.0269}$ & $\mathbf{0.2008 \pm 0.0261}$ \\ 
    & Oracle     & \textit{0.8446 ± 0.0065} & \textit{0.0768 ± 0.0125} \\ 
    \midrule
\multirow{4}{*}{Checkerboard} 
    & NPEB       & $0.5178 \pm 0.0311$ & $0.1866 \pm 0.0066$ \\
    & XDGMM      & $0.5101 \pm 0.0493$ & $0.1834 \pm 0.0233$ \\
    & MEC-MMD  & $\mathbf{0.4039 \pm 0.0116}$ & $\mathbf{0.1505 \pm 0.0103}$ \\ 
    & Oracle     & \textit{0.3958 ± 0.0057} & \textit{0.0700 ± 0.0044} \\ \bottomrule
\end{tabular}
}
\vspace{-4mm}
\end{wraptable}

Table~\ref{tab:detailed_results} reports MSE (between denoised estimates and latent ground truth) and Wasserstein-1 (W1) distance (between each method's learned prior and the test-set latent distribution). Figure \ref{fig:2d-exp} shows denoised scatter plots from a representative run. We observed that NPEB places the prior on a discrete support, producing clustered posterior means. XDGMM is smoother but is constrained by its Gaussian-mixture parameterization, leading to oversmoothing and poor recovery of sharp boundaries, particularly for Checkerboard, and the hollow geometry of Circle. Both baselines are further penalized by relying on a scalar estimate of noise rather than the true per-sample heteroscedastic levels. MEC-MMD sidesteps noise estimation entirely during training: the audit correction absorbs the bias introduced by measurement error without requiring its parametric form. It faithfully recovers the continuous geometry of all three manifolds across seeds, achieves the best MSE and W1 among non-oracle methods on every dataset, and remains robust to outlier contamination, consistent with MMD's robustness to small amounts of corrupted mass \cite{cherief2020mmd, nafshi2024partial}. On Moons and Checkerboard, MEC-MMD nearly closes the gap to the Oracle, demonstrating that a modest audit sample of 300 pairs is sufficient to recover near-optimal denoising even when the full noise distribution is withheld.

\begin{wrapfigure}{l}{0.47\linewidth}
    \centering
    \vspace{-5mm}
    \includegraphics[width=0.97\linewidth]{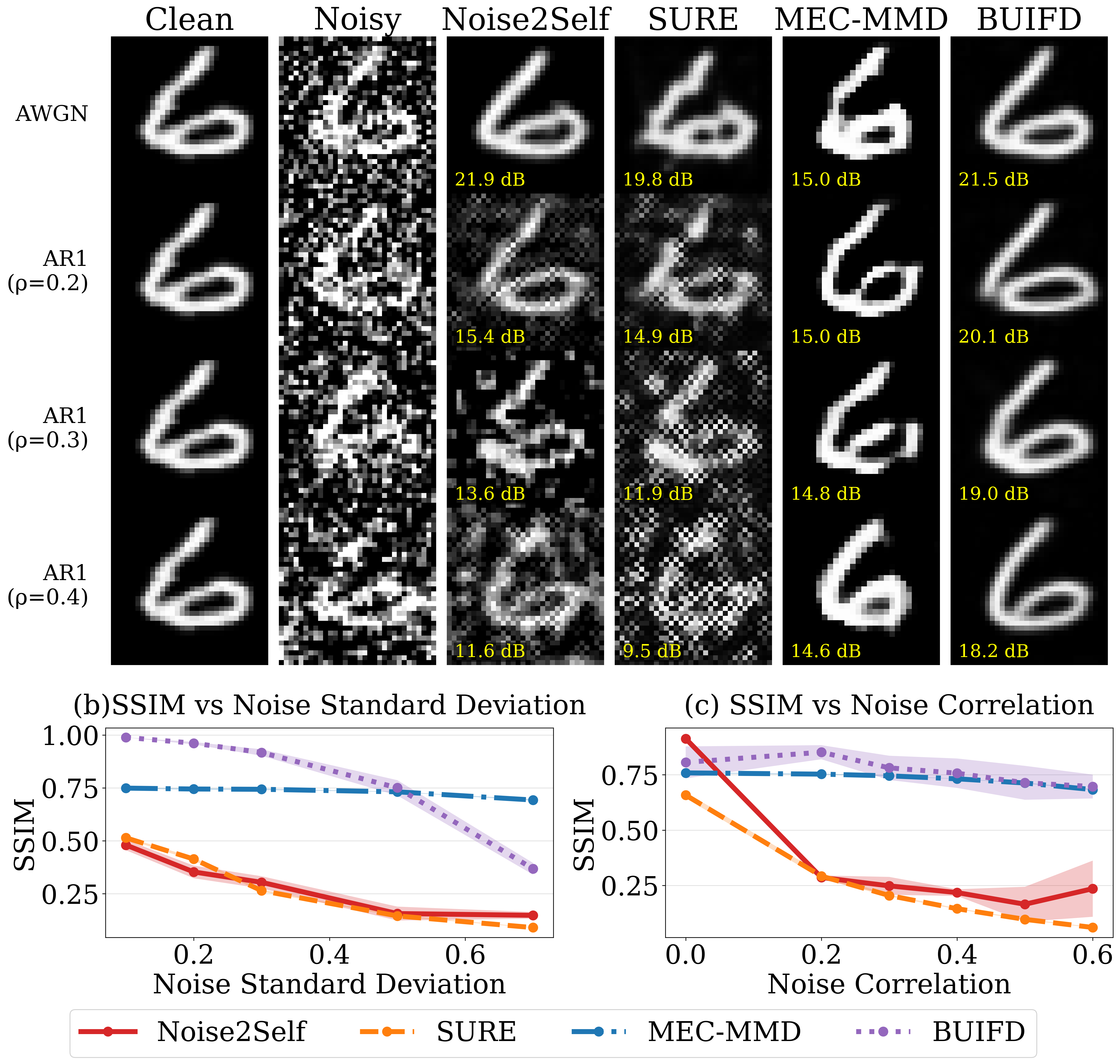}
    \vspace{-2mm}
    \caption{Image denoising performance.}
    \label{fig:mnist_comparison}
    \vspace{-5mm}
\end{wrapfigure}

 \vspace{-3mm}
\paragraph{Experiment 6 -- Robustness in Image Denoising.} While the method is not explicitly designed and optimized for image denoising, we demonstrate its applicability for unsupervised image denoising. The observed MNIST dataset contains only corrupted images, $\tilde{x}=x+u$, with known noise law $p(u)$, together with a small audit set of paired clean/noisy examples. We train a generator $G_\theta(z)$ as a clean-image prior by simulating the forward corruption process on generated samples and minimizing MEC-MMD between generated clean images and the observed noisy dataset. The learned prior is then used for empirical Bayes denoising by approximating $\mathbb{E}[x\mid \tilde{x}]$ with importance sampling on the generative manifold. We test on additive white Gaussian noise with $\sigma=0.5$ and spatially correlated Gaussian noise with the same total variance. Figure~\ref{fig:mnist_comparison} compares MEC-MMD with Noise2Self \cite{batson2019noise2self}, BUIFD \cite{el2020blind}, and SURE \cite{luisier2007new}. BUIFD and Noise2Self perform better under white noise but degrade as spatial correlation increases. MEC-MMD remains stable across correlation and noise levels, indicating that our method yields more robust denoising in the presence of structured measurement error. See limitations for discussion. 

\vspace{-3mm}
\paragraph{Experiment 7 -- Scaling with dimension.} We investigate the efficacy of our method in high dimensions by considering a deconvolution task on a one-dimensional spiral embedded in $\mathbb R^d$, $d\in\{3,7,11,15,20,30\}$, with $N=2000$ noisy observations and heteroscedastic Gaussian noise. All MMD-based methods fit the same normalizing flow with a Laplace kernel. All noise-aware methods share the same audit of $m=400$ points, and none is given the true noise level. convMMD resamples the audited residuals and XDGMM uses the audited residual variance, so the comparison is like-for-like in the unknown-noise setting. We measure performance by using the sliced Wasserstein distance on held-out latent points multiplied by $\sqrt d$ (Table~\ref{tab:dimension}; details in Appendix~\ref{app:details_experiments}). Between $d=3$ and $d=30$, naive MMD, convMMD and XDGMM degrade from $0.048$ to $0.208$, from $0.031$ to $0.062$ and from $0.030$ to $0.087$, whereas MEC-MMD stays essentially flat ($0.029$ to $0.034$). At $d=30$, MEC-MMD is about $1.8\times$ and $2.6\times$ better than convMMD and XDGMM, respectively. A plausible explanation is that MEC-MMD never models the noise and it corrects only the scalar kernel expectations that define $\mathrm{MMD}^2$, whereas others must estimate quantities that grow with $d$ (a $d\times d$ noise covariance or the $d$-dimensional residual distribution). 
 \vspace{-3mm}
\begin{table}[h]
\centering
\caption{Sliced Wasserstein distance $\times\sqrt d$ (lower is better) on the tangled-ribbon deconvolution task across dimensions $d$; $N=2000$, one audit of size $m=400$ shared by all noise-aware methods. Entries are mean $\pm$ standard error over 5 seeds. No method is given the true noise level.}
\label{tab:dimension}
\resizebox{\textwidth}{!}{%
\begin{tabular}{lcccccc}
\toprule
Method & $d=3$ & $d=7$ & $d=11$ & $d=15$ & $d=20$ & $d=30$ \\
\midrule
Na\"ive MMD    & $0.048 \pm 0.002$ & $0.071 \pm 0.001$ & $0.098 \pm 0.002$ & $0.123 \pm 0.001$ & $0.151 \pm 0.002$ & $0.208 \pm 0.002$ \\
convMMD        & $0.031 \pm 0.002$ & $\mathbf{0.031 \pm 0.001}$ & $0.037 \pm 0.003$ & $0.045 \pm 0.002$ & $0.047 \pm 0.001$ & $0.062 \pm 0.001$ \\
XDGMM          & $0.030 \pm 0.003$ & $0.033 \pm 0.001$ & $0.043 \pm 0.001$ & $0.053 \pm 0.001$ & $0.061 \pm 0.002$ & $0.087 \pm 0.001$ \\
MEC-MMD (Ours) & $\mathbf{0.029 \pm 0.005}$ & $0.033 \pm 0.006$ & $\mathbf{0.035 \pm 0.005}$ & $\mathbf{0.030 \pm 0.002}$ & $\mathbf{0.034 \pm 0.003}$ & $\mathbf{0.034 \pm 0.006}$ \\
\bottomrule
\end{tabular}%
}
\end{table}
\vspace{-2mm}

\vspace{-3mm}
\section{Discussion} \label{sec:discussion} \vspace{-2mm}

\paragraph{Bias--Variance Trade-off.} Although the proposed correction reduces MMD bias, it is not uniformly optimal because it trades bias reduction for variance inflation. Measurement error induces systematic bias in kernel evaluations and empirical MMD estimates; the audit-based Horvitz--Thompson correction removes this bias using the audited subset. However, because the audit sample is small, the correction adds sampling variability (Theorem~\ref{thm:me_variance_compact}) and can have higher variance than the naive estimator. Thus, when ME bias is small relative to correction variance, the naive biased estimator may achieve lower MSE, as in the left panel of Figure~\ref{fig:me_main_result} for $m<40$. Similarly, the audit-only estimator can outperform the combined estimator when the audit sample is sufficiently informative, or the noisy sample has low SNR, as in the middle panel of Figure~\ref{fig:me_main_result}. Thus, the proposed estimator is best viewed as part of an estimator-selection problem rather than a universally optimal choice. The preferred estimator depends on the relative sizes of ME bias, audit-sample variance, and correction-induced variance. In practice, cross-validation or sample splitting can compare the naive, audit-only, and corrected estimators and select the one with the lowest predictive or resampling-based risk. \vspace{-2mm}

\paragraph{Non-uniform Audit Sample.} We studied settings where an \textit{audit} or \textit{validation} sample can be obtained through simple random sampling. However, in many practical scenarios, audit samples are collected through non-random mechanisms, such as stratified, outcome-dependent, or uncertainty-guided validation designs~\cite{amorim2021two,nab2021sampling}. Section~\ref{sec:nonuniform_main} and Appendix~\ref{sec:nonuniform-audit} extend MEC-MMD with inverse-probability weights, and Experiment~4b shows that, when the inclusion probabilities are known, MEC-MMD performs well. While this extension preserves the basic kernel-level correction principle, it introduces additional methodological challenges, including estimating audit inclusion probabilities, controlling variance inflation from large weights, and developing stable or doubly robust alternatives. We therefore view non-uniform audit sampling as an important direction for future work. \vspace{-2mm}

\paragraph{Beyond Additive ME.}
Although the main formulation and experimental results are presented under an additive ME model, the audit-based correction is not intrinsically tied to additivity. The key requirement is that, for audited units, both the proxy observation {\small$\widetilde{Y}_j$} and the corresponding latent value $Y_j$ are observed, such that {\small $\Gamma_{jj'} = k(Y_j,Y_{j'}) - k(\widetilde{Y}_j,\widetilde{Y}_{j'})$} and  {\small$\Lambda_{ij} = k(X_i,Y_j) - k(X_i,\widetilde{Y}_j)$} can be computed. Thus, the same correction principle can in principle be applied to multiplicative noise, rounding, nonlinear sensor distortion, or other proxy mechanisms. However, this statistical generality should be distinguished from formal identifiability. The audit sample must be representative, or else properly weighted; each proxy observation should correspond to a well-defined latent unit; and identifiability of the latent target distribution under the measurement mechanism must be established separately. In settings involving many-to-one coarsening, censoring or truncation, selection-biased audits, or extremely low signal-to-noise ratios, MEC-MMD may remain well-defined but can suffer from high variance or may fail to recover the desired latent target without additional assumptions. \vspace{-2mm}

\paragraph{Limitations and Open Challenges.} First, the proposed method has not been systematically evaluated how the required audit sample size should scale with dimension and this remains an open question. Second, the current approach does not explicitly exploit structural information about the ME mechanism when such information is available. The method is intentionally designed for a \textit{distribution-free noise setting}, where the analyst may not know the ME distribution. Although this robustness is attractive, it necessarily leaves potentially informative structure unused. In settings where the noise distribution is known, can be consistently estimated, or sampled \cite[e.g.,][]{thurow2025characterizing, glazer2025beyond, farahi2019mass, chen2019calibration, gebhardt2021hobby, wei2020physics}, methods that model the measurement process directly may be preferable \cite[e.g.,][]{vashistha2026convolutional} and achieve potentially a lower variance and greater statistical efficiency. A further open question concerns kernel choice and bandwidth selection. In particular, there is no clear reason to expect that the optimal kernel bandwidth for the clean sample appearing in the audit-correction term should coincide with the bandwidth used elsewhere in the estimator. This suggests that kernel optimization, carried out with appropriate care, may provide an additional avenue for improving performance. \vspace{-1mm}

Although we include an image denoising case study, it is not the primary focus of this work. Image denoising is a mature field with specialized benchmarks, methods, and noise models \cite[e.g., see,][]{dabov2007image, buades2005review, zhang2023hyperspectral, elad2023image, izadi2023image}. Our method is designed for general measurement-error correction and does not exploit image-specific structure or common imaging-noise assumptions. Thus, although we showed that MEC-MMD is robust to the noise structure, we do not claim it is optimized for image denoising. We therefore leave image-specific extensions and optimization to future work.
\vspace{-2mm}

\paragraph{Contributions.} We formulate MMD-based distribution matching under unknown ME with abundant noisy data and a small audit sample. We propose MEC-MMD, which corrects kernel evaluations to target the latent rather than proxy distribution. We prove design-unbiasedness, derive its variance, establish consistency, and demonstrate empirical gains in estimation, density learning, and denoising. Our results provide a practical foundation for robust distribution matching in noisy settings. %
 \vspace{-2mm}

\begin{ack}
The authors acknowledge support from the National Science Foundation under Cooperative Agreement No. 2421782 and from the Simons Foundation under Award No. MPS-AI-00010515, through the NSF–Simons AI Institute for Cosmic Origins (CosmicAI, \url{https://cosmicai.org/}). Additionally, the authors acknowledge the Texas Advanced Computing Center (TACC, \url{http://www.tacc.utexas.edu}) at The University of Texas at Austin for providing computational resources that have contributed to the research results reported within this paper. 
\end{ack}

{
\small

\bibliographystyle{unsrt}
\bibliography{bibliography}%

}

\newpage

\appendix

\section*{Positioning}
\label{sec:related_work}

We consider the problem of parameter estimation with an MMD loss when the scientific target is observed only through noisy measurements, together with a small audited subset of clean latent values. This problem sits at the intersection of kernel mean embeddings and MMD-based estimation, simulation-based inference, and classical measurement-error methodology.

\paragraph{Noisy and Gold-Standard Data Regimes in Scientific Practice.} Modern scientific workflows frequently operate in a regime where two distinct data sources coexist: a large-scale, inexpensive, and error-prone measurement process, and a smaller, high-quality gold-standard measurement. The scalable data source enables broad coverage and statistical power, arising from mechanisms such as automated labeling, crowd-sourcing, proxy variables, passive data collection, or low-cost instrumentation. In contrast, gold-standard data are typically obtained through expert annotation, controlled experimental procedures, or high-precision instruments, and are therefore substantially more costly and time-intensive to acquire.

\begin{table}[h!]
\centering
\small
\begin{adjustbox}{width=\textwidth}
\begin{tabular}{lll}
\hline
\textbf{Domain} & \textbf{Noisy / Scalable Measurement} & \textbf{Gold-Standard Measurement} \\
\hline
Natural Language Processing & Heuristic or LLM-generated labels & Expert manual annotation \\
Citizen Science (Astronomy/Ecology) & Crowd-sourced labels or reports & Expert-reviewed classifications \\
Computer Vision & Web-scraped labels & Curated human annotations \\
Speech Processing & Automatically transcribed audio (ASR) & Human-transcribed speech \\
Medicine / Epidemiology & Proxy variables (e.g., ICD codes, questionnaires) & Clinical chart review or biomarkers \\
Genomics & High-throughput sequencing (noisy reads) & Sanger sequencing / validated assays \\
Astronomy / Physics & Photometric imaging & Spectroscopic measurements \\
Environmental Monitoring & Low-cost sensors & Regulatory-grade instruments \\
Remote Sensing & Satellite-derived proxies & Ground-truth field measurements \\
Economics / Social Science & Survey self-reports or administrative proxies & Audited records / verified outcomes \\
Recommender Systems & Implicit feedback (clicks, views) & Explicit ratings or curated preferences \\
Education Research & Automated grading / engagement logs & Expert grading or standardized assessments \\
\hline
 \vspace{1mm}
\end{tabular}
\end{adjustbox}
\caption{Scientific domains exhibiting a combination of abundant noisy measurements and sparse gold-standard observations.}
\label{tab:noisy_gold_patterns}
\end{table}

As a result of this cost asymmetry, gold-standard measurements are typically available only for a subset of observations, forming what is often referred to as a validation or audit sample. This pattern arises consistently across disciplines (see Table~\ref{tab:noisy_gold_patterns} for examples). In machine learning, for example, large datasets are often labeled using heuristics, weak supervision, or automated systems, with a smaller subset manually verified. In the natural sciences, high-throughput but noisy measurement technologies (e.g., remote sensing, imaging, or sequencing) are complemented by precise but lower-throughput methods like spectroscopy, field sampling, or targeted assays. In the social sciences and economics, self-reported or administrative data are frequently validated against audited records or carefully curated ground-truth datasets.

This setting represents a fundamental trade-off between scale and accuracy. Noisy measurements provide breadth but may introduce systematic bias, while gold-standard data provide accuracy but are limited in scope. Consequently, many modern empirical pipelines are designed to integrate these complementary sources, leveraging gold-standard subsets to calibrate, validate, or correct large-scale noisy datasets. The prevalence of this paradigm across fields suggests that it constitutes a general and unifying framework for understanding data quality, experimental design, and inference under imperfect measurement.

\paragraph{MMD as a Discrepancy for Estimation and SBI.}
Kernel mean embeddings provide a flexible nonparametric representation of distributions in RKHSs, and MMD is the induced discrepancy obtained by comparing these embeddings \citep{muandet2017kernel,gretton2012kernel}. Beyond two-sample testing, MMD is now widely used as an estimation objective because it avoids explicit density estimation and does not require tractable likelihoods. This has made it attractive in generative modeling, minimum-distance estimation, and simulation-based inference \citep{li2015generative,briol2019statistical,cranmer2020frontier,lueckmann2021benchmarking}. MMD has been widely used in hypothesis testing \citep{gretton2012kernel}, generative modeling \citep{li2017mmd}, domain adaptation \citep{long2017deep}, and simulation-based inference \citep{briol2019statistical,bharti2023optimally}. In SBI specifically, kernel and MMD-based discrepancies have been used in approximate Bayesian computation, robust generalized Bayesian procedures, and optimally weighted likelihood-free estimators \citep{park2016k2,dellaporta2022robust,bharti2023optimally}. However, this literature typically assumes that the observed sample is itself the inferential target. In our setting, the observed sample is a noisy proxy for a latent target distribution, so the naive empirical MMD loss is systematically biased toward the measurement process.

\paragraph{Measurement Error and Latent-variable Correction.}
Classical measurement-error methods address this mismatch through deconvolution, errors-in-variables regression, SIMEX, and likelihood-based latent-variable modeling \cite{fan1991optimal,cook1994simulation,carroll2006measurement,kelly2007some}. These approaches provide powerful tools for density estimation and regression under noisy observation, but they often rely on strong assumptions on the noise law, parametric structure, replicate measurements, or identifiability conditions, and many become unstable in high-dimensional or weakly supervised settings. More importantly for our purposes, they are usually designed to recover the latent distribution or regression function directly, rather than to correct a distribution-matching loss used inside a simulator-based estimation pipeline. Our approach is different: rather than solving the full deconvolution problem, we correct the MMD objective itself so that optimization remains targeted at the latent scientific quantity.

\paragraph{Audit-based Correction.}
A growing line of work, such a prediction-powered inference paradigm \citep{angelopoulos2023prediction,zrnic2024cross,datta2025prediction,song2026demystifying,kluger2026m}, studies how to combine a large sample of inexpensive proxy measurements with a smaller set of gold-standard observations to enable valid inference. These approaches exploit a simple but powerful principle: the proxy data provide statistical efficiency and coverage, while the audited sample supplies a correction signal that mitigates bias induced by measurement error or model misspecification. Importantly, they avoid requiring the proxy to be fully accurate, instead using the gold-standard subset to adjust downstream estimators in a data-driven manner.

Existing methods in this framework focus primarily on low-dimensional targets such as means, quantiles, or regression parameters, where corrections can be expressed through bias-adjusted estimating equations or reweighting schemes. In contrast, our setting involves a distributional objective (MMD) defined through expectations of kernel features. This introduces additional challenges, as the discrepancy is nonlinear in the data-generating distribution and is evaluated implicitly through pairwise comparisons.

Our approach extends the audit-based paradigm to this setting by constructing corrected estimators of the kernel expectations that define MMD. Rather than modifying the downstream estimator after the fact, we directly adjust the loss function used for estimation. This preserves the computational and modeling advantages of MMD-based procedures while ensuring that optimization remains targeted at the latent distribution of scientific interest, rather than the noisy proxy observations.

\paragraph{Positioning.}
The existing literatures therefore provide only part of the solution. MMD-based SBI supplies a flexible likelihood-free objective but generally assumes clean data; classical measurement-error methods address noisy observations but are not tailored to kernel distribution-matching losses; and audit-based inference provides a correction template but has not been developed for MMD-based parameter estimation with latent targets. Our contribution is to combine these strands: we use MMD as the estimation loss, target latent density and regression parameters under measurement error, and exploit a small audited subset to debias the loss without requiring full deconvolution or exact specification of the measurement process.

\section{Assumptions}\label{sec:assumptions}
We restate Assumptions ~\ref{ass:bounded_kernel} - ~\ref{ass:asymptotic_regime} from the main text here for completness, along with additional assumptions.
\begin{assumption}[Additive measurement-error generative model]
\label{ass:additive_me_model}
The latent target sample $Y_1,\dots,Y_N$ is drawn i.i.d.\ from $q$, and for each $j\in\{1,\dots,N\}$,
\[
\phi_j \overset{iid}{\sim} \Pi,
\qquad
\varepsilon_j \mid \phi_j \sim P(\cdot \mid \phi_j),
\qquad
\widetilde Y_j = Y_j + \varepsilon_j.
\]
The triples $(Y_j,\phi_j,\varepsilon_j)$ are independent across $j$, and are independent of the clean reference sample $X_1,\dots,X_n \overset{iid}{\sim} p$.
\end{assumption}

\begin{assumption}[Simple random auditing]
\label{ass:srs_audit_app}
The audit set $L \subset \{1,\dots,N\}$ is a simple random sample without replacement of fixed size $m$ with $2\le m\le N$, drawn independently of the data-generating process. In particular,
\[
\Pr(j\in L)=\frac{m}{N},
\qquad
\Pr(j,j'\in L)=\frac{m(m-1)}{N(N-1)}
\quad\text{for } j\neq j'.
\]
\end{assumption}

\begin{assumption}[Asymptotic regime]
\label{ass:asymptotic_regime_app}
We consider a sequence of problems with $n,N,m\to\infty$ such that
\[
\frac{m}{N}\to \rho \in (0,1],
\qquad
\frac{n}{N}\to \gamma \in (0,\infty).
\]
\end{assumption}

\begin{assumption}[Bounded kernel]  \label{ass:bounded_kernel_app}
There exists $K<\infty$ such that
\[
\sup_{x,x'\in\mathcal X} |k(x,x')| \le K \quad \forall x, x' \in \mathcal{X}.
\]
\end{assumption}

\subsection{Notation}
\label{subsec:notation}

Table~\ref{tab:notation} summarizes the notation used throughout the paper. We group symbols into four categories: (i) probabilistic objects, (ii) RKHS and MMD-related quantities, (iii) sampling and measurement-error components, and (iv) estimator-specific terms. This table is intended as a reference guide; we introduce symbols locally where needed, but maintain consistency with the definitions summarized here.

\begin{table*}[th!]
\centering
\caption{Summary of notation.}
\label{tab:notation}
\scriptsize
\setlength{\tabcolsep}{3pt}
\renewcommand{\arraystretch}{1.1}
\resizebox{\textwidth}{!}{
\begin{tabular}{llll}
\toprule
Notation & Type & Status & Description \\
\midrule
$\mathcal X\subseteq\mathbb R^d$ & Set & Constant & Sample space. \\
$d$ & Scalar & Constant & Ambient dimension. \\
$p,q$ & Distribution & Constant & Reference and latent target distributions. \\
$X,X'$ & Vector & Random variable & iid draws from $p$. \\
$Y,Y'$ & Vector & Random variable & iid latent draws from $q$. \\
$x,y$ & Vector & Realization & Realized values of $X,Y$. \\
$\widetilde Y$ & Vector & Random variable & Noisy proxy for $Y$. \\
$\varepsilon$ & Vector & Random variable & Additive measurement error. \\
$Z$ & Vector & Random variable & Covariate affecting noise. \\
$\phi$ & Vector/other & Random variable & Unit-specific noise parameter. \\
$P(\cdot\mid\phi)$ & Distribution & Model & Conditional noise distribution. \\
$\Pi$ & Distribution & Model & Distribution over noise parameters. \\
\midrule
$\mathcal H$ & Function space & Constant & RKHS induced by kernel $k$. \\
$f$ & Function & Variable & RKHS test function. \\
$k$ & Function & Constant & Positive definite kernel. \\
$\mathrm{MMD}^2(p,q;k)$ & Scalar & Target & Squared MMD. \\
\midrule
$n,N,m$ & Scalars & Constants & Sample sizes (clean, target, audit). \\
$i,j,j'$ & Indices & Deterministic & Observation indices. \\
$X_i$ & Vector & Random variable & Clean reference observation. \\
$Y_j$ & Vector & Random variable & Latent target observation. \\
$\widetilde Y_j$ & Vector & Random variable & Noisy target proxy. \\
$\varepsilon_j$ & Vector & Random variable & Measurement error. \\
$L$ & Set & Random set & Audited indices. \\
$\mathbb E_L,\mathbb P_L$ & Operators & Design-based & Expectation/probability over audit design. \\
\midrule
$A_{pp}^\star$ & Scalar & Oracle statistic & Clean--clean kernel sum. \\
$A_{qq}^\star$ & Scalar & Oracle statistic & Latent--latent kernel sum. \\
$A_{pq}^\star$ & Scalar & Oracle statistic & Clean--latent kernel sum. \\
$\widehat{\mathrm{MMD}}_\star^2$ & Scalar & Oracle estimator & Empirical MMD using latent data. \\
$\Gamma_{jj'}$ & Scalar & Realization & Correction for $qq$ kernel pair. \\
$\Lambda_{ij}$ & Scalar & Realization & Correction for $pq$ kernel pair. \\
$\widehat A_{qq}$ & Scalar & Estimator & Corrected latent--latent sum. \\
$\widehat A_{pq}$ & Scalar & Estimator & Corrected clean--latent sum. \\
$\widehat{\mathrm{MMD}}^2_{\mathrm{MEC}}$ & Scalar & Estimator & Measurement-error-corrected MMD. \\
\midrule
$\Lambda_j$ & Scalar & Realization & Unit-level $pq$ correction. \\
$\bar\Lambda$ & Scalar & Realization & Mean of $\Lambda_j$. \\
$\bar\Gamma$ & Scalar & Realization & Mean of $\Gamma_{jj'}$. \\
$\bar\Gamma_j$ & Scalar & Realization & Unit-level $qq$ correction. \\
$S_{\mathrm{proj}}^2$ & Scalar & Realization & First-order projection variance. \\
$R_{\mathrm{deg}}^2$ & Scalar & Realization & Second-order remainder. \\
$\rho$ & Scalar & Limit & Audit fraction $m/N$. \\
$V_{\mathrm{design}}$ & Scalar & Limit & Asymptotic design variance. \\
\bottomrule
\end{tabular}
}
\end{table*}

\newpage

\section{Proofs}

\begin{lemma} \label{lem:me_ht_identities}
Fix any finite population and let $L$ be a simple random sample without replacement of size $m$ from $\{1,\dots,N\}$. For any real-valued function $g:\{1,\dots,N\}\to\mathbb R$,
\begin{equation}
\label{eq:me_ht_single}
\mathbb E_L\!\left[\frac{N}{m}\sum_{j\in L} g(j)\right] = \sum_{j=1}^N g(j).
\end{equation}
For any real-valued function $h:\{1,\dots,N\}^2\to\mathbb R$ with $h(j,j)=0$,
\begin{equation}
\label{eq:me_ht_pair}
\mathbb E_L\!\left[ \frac{N(N-1)}{m(m-1)} \sum_{\substack{j,j'\in L\\ j\neq j'}} h(j,j') \right] = \sum_{\substack{j,j'=1\\ j\neq j'}}^N h(j,j').
\end{equation}
\end{lemma}

\begin{proof}[Proof of Lemma~\ref{lem:me_ht_identities}.]
Let
\[
I_j := \mathbf{1}\{j\in L\},
\qquad j=1,\dots,N,
\]
denote the audit inclusion indicator for unit $j$. Under simple random sampling without replacement of size $m$,
\[
\mathbb E_L[I_j] = \Pr(j\in L)=\frac{m}{N}.
\]
Likewise, for distinct $j\neq j'$,
\[
\mathbb E_L[I_j I_{j'}]
=
\Pr(j,j'\in L)
=
\frac{m(m-1)}{N(N-1)}.
\]

We prove the two identities separately.

Single-index identity. Write
\[
\sum_{j\in L} g(j) = \sum_{j=1}^N I_j\, g(j).
\]
Therefore, by linearity of expectation,
\begin{align*}
\mathbb E_L\!\left[\frac{N}{m}\sum_{j\in L} g(j)\right]
&=
\frac{N}{m}\, \mathbb E_L\!\left[\sum_{j=1}^N I_j g(j)\right] \\
&=
\frac{N}{m}\sum_{j=1}^N g(j)\,\mathbb E_L[I_j] \\
&=
\frac{N}{m}\sum_{j=1}^N g(j)\,\frac{m}{N} \\
&=
\sum_{j=1}^N g(j),
\end{align*}
which proves the claim.

\textit{Pairwise identity.}
Assume $m\ge 2$, so that the normalizing factor $m(m-1)$ is well defined. Since $h(j,j)=0$, only off-diagonal pairs matter, and we may write
\[
\sum_{\substack{j,j'\in L\\ j\neq j'}} h(j,j') = \sum_{\substack{j,j'=1\\ j\neq j'}}^N I_j I_{j'}\, h(j,j').
\]
Taking expectation and using linearity,
\begin{align*}
\mathbb E_L\!\left[ \frac{N(N-1)}{m(m-1)} \sum_{\substack{j,j'\in L\\ j\neq j'}} h(j,j') \right]
&=
\frac{N(N-1)}{m(m-1)} \mathbb E_L\!\left[ \sum_{\substack{j,j'=1\\ j\neq j'}}^N I_j I_{j'}\, h(j,j') \right] \\
&=
\frac{N(N-1)}{m(m-1)} \sum_{\substack{j,j'=1\\ j\neq j'}}^N h(j,j')\,\mathbb E_L[I_j I_{j'}].
\end{align*}
For each $j\neq j'$,
\[
\mathbb E_L[I_j I_{j'}] = \Pr(j,j'\in L) = \frac{m(m-1)}{N(N-1)}.
\]
Substituting this into the previous display gives
\begin{align*}
\mathbb E_L\!\left[ \frac{N(N-1)}{m(m-1)} \sum_{\substack{j,j'\in L\\ j\neq j'}} h(j,j') \right]
&=
\frac{N(N-1)}{m(m-1)} \sum_{\substack{j,j'=1\\ j\neq j'}}^N
h(j,j')\, \frac{m(m-1)}{N(N-1)} \\
&=
\sum_{\substack{j,j'=1\\ j\neq j'}}^N h(j,j'),
\end{align*}
which proves the claim.
\end{proof}

\begin{proof}[Proof of Theorem~\ref{thm:me_unbiased_sums}.]
For the $pq$ term, write
\[
A_{pq}^\star = \sum_{i=1}^n\sum_{j=1}^N k(X_i,Y_j)
= \sum_{i=1}^n\sum_{j=1}^N k(X_i,\widetilde Y_j) +
\sum_{j=1}^N g(j),
\]
where
\[
g(j) := \sum_{i=1}^n
\Big(
k(X_i,Y_j)-k(X_i,\widetilde Y_j)
\Big)
= \sum_{i=1}^n \Lambda_{ij}.
\]
By Lemma~\ref{lem:me_ht_identities},
\[
\mathbb E_L\!\left[
\frac Nm \sum_{j\in L} g(j)
\right]
=
\sum_{j=1}^N g(j),
\]
which implies $\mathbb E_L[\widehat A_{pq}] = A_{pq}^\star$.

For the $qq$ term, write
\[
A_{qq}^\star
=
\sum_{j\neq j'} k(Y_j,Y_{j'})
=
\sum_{j\neq j'} k(\widetilde Y_j,\widetilde Y_{j'})
+
\sum_{j\neq j'} h(j,j'),
\]
where
\[
h(j,j')
:=
k(Y_j,Y_{j'})
-
k(\widetilde Y_j,\widetilde Y_{j'})
=
\Gamma_{jj'},
\qquad h(j,j):=0.
\]
Again by Lemma~\ref{lem:me_ht_identities},
\[
\mathbb E_L\!\left[ \frac{N(N-1)}{m(m-1)} \sum_{\substack{j,j'\in L\\ j\neq j'}} h(j,j') \right] = \sum_{j\neq j'} h(j,j'),
\]
so $\mathbb E_L[\widehat A_{qq}] = A_{qq}^\star$.

Finally, $A_{pp}^\star$, $n(n-1)$, $N(N-1)$, and $nN$ are non-random under the audit design, so taking expectations term-by-term in \eqref{eq:mmd_me} yields
\[
\mathbb E_L[\widehat{\mathrm{MMD}}^2_{\mathrm{MEC}}] = \widehat{\mathrm{MMD}}_\star^2.
\]
\end{proof}

\begin{proof}[Proof of Theorem~\ref{thm:me_variance_compact}.]
Starting from \eqref{eq:Aqq_me_hat}, \eqref{eq:Apq_me_hat}, and \eqref{eq:mmd_me},
\begin{equation}
\label{eq:proof_start_compact}
\widehat{\mathrm{MMD}}_{\mathrm{MEC}}^2-\widehat{\mathrm{MMD}}_\star^2
=
\left[
\binom{m}{2}^{-1}
\sum_{\substack{1\le j<j'\le N\\ j,j'\in L}}
\Gamma_{jj'}-\bar\Gamma
\right]
-
2\left[
\frac1m\sum_{j\in L}\Lambda_j-\bar\Lambda
\right].
\end{equation}

Now decompose the finite-population kernel residuals as
\[
\Gamma_{jj'}
=
\bar\Gamma+\alpha_j+\alpha_{j'}+\beta_{jj'},
\qquad 1\le j<j'\le N,
\]
where
\[
\alpha_j:=\frac{N-1}{N-2}(\bar\Gamma_j-\bar\Gamma),
\qquad
\beta_{jj'}
:=
\Gamma_{jj'}-\bar\Gamma-\alpha_j-\alpha_{j'}.
\]
By construction,
\[
\sum_{j=1}^N \alpha_j=0,
\qquad
\sum_{j'\neq j}\beta_{j\wedge j',\,j\vee j'}=0
\quad\text{for every }j.
\]
Substituting this decomposition into \eqref{eq:proof_start_compact} gives
\[
\widehat{\mathrm{MMD}}_{\mathrm{MEC}}^2-\widehat{\mathrm{MMD}}_\star^2
=
\frac{2}{m}\sum_{j\in L}\bigl[\alpha_j-(\Lambda_j-\bar\Lambda)\bigr]
+
\binom{m}{2}^{-1}
\sum_{\substack{1\le j<j'\le N\\ j,j'\in L}}\beta_{jj'}.
\]
Define
\[
u_j:=\alpha_j-(\Lambda_j-\bar\Lambda),
\qquad
V_L:=\binom{m}{2}^{-1}
\sum_{\substack{1\le j<j'\le N\\ j,j'\in L}}\beta_{jj'}.
\]
Then
\begin{equation}
\label{eq:proj_deg_split_compact}
\widehat{\mathrm{MMD}}_{\mathrm{MEC}}^2-\widehat{\mathrm{MMD}}_\star^2
=
\frac{2}{m}\sum_{j\in L}u_j+V_L.
\end{equation}
Moreover, $\sum_{j=1}^N u_j=0$, and by definition
\[
S_{\mathrm{proj}}^2=\frac1{N-1}\sum_{j=1}^N u_j^2,
\qquad
R_{\mathrm{deg}}^2=\binom{N}{2}^{-1}\sum_{1\le j<j'\le N}\beta_{jj'}^2.
\]
By Assumption~\ref{ass:bounded_kernel} and the triangle inequality (each of $\Lambda_{ij},\Gamma_{jj'}$ is a difference of two kernel values, each bounded by $K$),
\begin{equation}\label{eq:corrections_bounded}
|\Lambda_{ij}|\le 2K,\qquad |\Gamma_{jj'}|\le 2K \qquad\text{for all } i,\ j\neq j'.
\end{equation}
The variance of the linear term is the standard finite-population sampling variance:
\[
\Var_L\!\left(\frac1m\sum_{j\in L}u_j\right)
=
\frac{N-m}{mN}\,S_{\mathrm{proj}}^2.
\]
Hence
\begin{equation}
\label{eq:var_linear_compact}
\Var_L\!\left(\frac{2}{m}\sum_{j\in L}u_j\right)
=
\frac{4(N-m)}{mN}\,S_{\mathrm{proj}}^2.
\end{equation}

For the degenerate term, the row-sum condition
\[
\sum_{j'\neq j}\beta_{j\wedge j',\,j\vee j'}=0
\]
implies the standard finite-population variance formula for a degenerate second-order U-statistic:
\begin{equation}
\label{eq:var_deg_compact}
\Var_L(V_L)
=
\frac{2(N-m)(N-m-1)}{m(m-1)(N-2)(N-3)}\,R_{\mathrm{deg}}^2.
\end{equation}
The same degeneracy implies orthogonality between the first-order and second-order parts,
\begin{equation}
\label{eq:cov_zero_compact}
\Cov_L\!\left(\frac{2}{m}\sum_{j\in L}u_j,\;V_L\right)=0.
\end{equation}
Combining \eqref{eq:proj_deg_split_compact}, \eqref{eq:var_linear_compact},
\eqref{eq:var_deg_compact}, and \eqref{eq:cov_zero_compact} yields
\eqref{eq:me_variance_compact}.
\end{proof}

\begin{proof}[Proof of Corollary~\ref{cor:me_variance_asymptotic}]
\emph{Boundedness.} By Assumption~\ref{ass:bounded_kernel} and \eqref{eq:corrections_bounded}, $|\Gamma_{jj'}|,|\Lambda_{ij}|\le 2K$, hence $|\bar\Gamma|,|\bar\Gamma_j|,|\Lambda_j|,|\bar\Lambda|\le 2K$. Let $\alpha_j,\beta_{jj'},u_j$ be as in the proof of Theorem~\ref{thm:me_variance_compact}. Since $\frac{N-1}{N-2}\le\frac32$ for $N\ge4$, $|\alpha_j|\le 6K$ and $|u_j|\le|\alpha_j|+|\Lambda_j-\bar\Lambda|\le 10K$, so
\[
S_{\mathrm{proj}}^2\le \tfrac{N}{N-1}\,100K^2\le 134K^2 .
\]
Moreover $\sum_j\alpha_j=0$ and $\sum_{j'\neq j}\beta_{j\wedge j',j\vee j'}=0$ for every $j$, so $\sum_{j<j'}\beta_{jj'}(\bar\Gamma+\alpha_j+\alpha_{j'})=0$. Hence $\sum_{j<j'}\Gamma_{jj'}^2=\sum_{j<j'}(\bar\Gamma+\alpha_j+\alpha_{j'})^2+\sum_{j<j'}\beta_{jj'}^2\ge\sum_{j<j'}\beta_{jj'}^2$, and
\[
R_{\mathrm{deg}}^2\le \binom{N}{2}^{-1}\sum_{j<j'}\Gamma_{jj'}^2\le 4K^2 .
\]
Thus $S_{\mathrm{proj}}^2$ and $R_{\mathrm{deg}}^2$ are $O(1)$ uniformly in $N$. 
By Theorem~\ref{thm:me_variance_compact},
\[
\Var_L\!\left(\widehat{\mathrm{MMD}}_{\mathrm{MEC}}^2\right)
=
\frac{4(N-m)}{mN}\,S_{\mathrm{proj}}^2
+
\frac{2(N-m)(N-m-1)}{m(m-1)(N-2)(N-3)}\,R_{\mathrm{deg}}^2.
\]
Since $m/N\to \rho\in(0,1]$,
\[
\frac{N-m}{mN}
=
\frac{1-m/N}{m}
=
\frac{1-\rho}{\rho N}+o(N^{-1}),
\]
and therefore
\[
\frac{4(N-m)}{mN}
=
\frac{4(1-\rho)}{\rho N}+o(N^{-1}).
\]
Similarly,
\[
\frac{(N-m)(N-m-1)}{m(m-1)(N-2)(N-3)}
=
\frac{(1-m/N)^2}{(m/N)^2\,N^2}+o(N^{-2})
=
\frac{(1-\rho)^2}{\rho^2 N^2}+o(N^{-2}),
\]
so
\[
\frac{2(N-m)(N-m-1)}{m(m-1)(N-2)(N-3)}
=
\frac{2(1-\rho)^2}{\rho^2 N^2}+o(N^{-2}).
\]
Substituting these two expansions into the exact variance formula yields
\[
\Var_L\!\left(\widehat{\mathrm{MMD}}_{\mathrm{MEC}}^2\right)
=
\left(
\frac{4(1-\rho)}{\rho N}+o(N^{-1})
\right)S_{\mathrm{proj}}^2
+
\left(
\frac{2(1-\rho)^2}{\rho^2 N^2}+o(N^{-2})
\right)R_{\mathrm{deg}}^2.
\]
Because $S_{\mathrm{proj}}^2=O(1)$ and $R_{\mathrm{deg}}^2=O(1)$ by the bounds above, the first remainder term is $o(N^{-1})$ and the second remainder term is $o(N^{-2})$, hence in particular $o(N^{-1})$. Therefore
\[
\Var_L\!\left(\widehat{\mathrm{MMD}}_{\mathrm{MEC}}^2\right)
=
\frac{4(1-\rho)}{\rho N}\,S_{\mathrm{proj}}^2
+
\frac{2(1-\rho)^2}{\rho^2 N^2}\,R_{\mathrm{deg}}^2
+
o(N^{-1}),
\]
which proves \eqref{eq:me_variance_asymptotic}.

The final display follows because the second term is $O(N^{-2})$ by the bounds above, which is of smaller order than the first term whenever $\rho<1$ and $N S_{\mathrm{proj}}^2\to\infty$ (e.g.\ $S_{\mathrm{proj}}^2$ bounded away from $0$) and therefore, the first-order approximation is
\[
\Var_L\!\left(\widehat{\mathrm{MMD}}_{\mathrm{MEC}}^2\right)
\approx
\frac{4(N-m)}{mN}\,S_{\mathrm{proj}}^2.
\qedhere
\]
\end{proof}

\begin{proof}[Proof of Theorem~\ref{thm:me_design_clt}.]
We first prove the rate statements.

For the $pq$ term, define
\[
g_j:=\frac1n\sum_{i=1}^n \Lambda_{ij} = \frac1n\sum_{i=1}^n \Bigl(k(X_i,Y_j)-k(X_i,\widetilde Y_j)\Bigr),
\qquad j=1,\dots,N.
\]
By Equation~\eqref{eq:Apq_me_hat},
\[
\widehat A_{pq}-A_{pq}^\star
=
N\left(\frac1m\sum_{j\in L} g_j-\frac1N\sum_{j=1}^N g_j\right)n,
\]
hence
\begin{equation}
\label{eq:pq_ht_error}
\frac{\widehat A_{pq}-A_{pq}^\star}{nN}
=
\frac1m\sum_{j\in L} g_j-\frac1N\sum_{j=1}^N g_j.
\end{equation}
Because Assumption~\ref{ass:bounded_kernel} gives
\[
|\Lambda_{ij}|
\le |k(X_i,Y_j)|+|k(X_i,\widetilde Y_j)|
\le 2K,
\]
we have $|g_j|\le 2K$ for every $j$. Under Assumption~\ref{ass:srs_audit}, the right-hand side of Equation~\eqref{eq:pq_ht_error} is the centered sample mean under simple random sampling without replacement from a bounded finite population. Its design variance is therefore
\[
\Var_L\!\left(
\frac{\widehat A_{pq}-A_{pq}^\star}{nN}
\right)
=
\frac{N-m}{mN}\,S_{g,N}^2,
\]
where
\[
S_{g,N}^2:=\frac1{N-1}\sum_{j=1}^N (g_j-\bar g_N)^2,
\qquad
\bar g_N:=\frac1N\sum_{j=1}^N g_j.
\]
Since $|g_j|\le 2K$, we have $S_{g,N}^2\le 4\frac{N}{N-1}K^2$, and Assumption~\ref{ass:asymptotic_regime} implies $(N-m)/(mN)=O(m^{-1})$. Hence
\[
\Var_L\!\left(
\frac{\widehat A_{pq}-A_{pq}^\star}{nN}
\right)
=
O(m^{-1}),
\]
which yields
\[
\frac{\widehat A_{pq}-A_{pq}^\star}{nN}
=
O_{\mathbb P_L}(m^{-1/2}).
\]

For the $qq$ term, define for $j<j'$,
\[
u_{jj'}
:=
\Gamma_{jj'}
=
k(Y_j,Y_{j'})-k(\widetilde Y_j,\widetilde Y_{j'}).
\]
Then by Equation~\eqref{eq:Aqq_me_hat},
\[
\frac{\widehat A_{qq}-A_{qq}^\star}{N(N-1)}
=
\binom{m}{2}^{-1}
\sum_{\substack{1\le j<j'\le N\\ j,j'\in L}} u_{jj'}
-
\binom{N}{2}^{-1}
\sum_{1\le j<j'\le N} u_{jj'}.
\]
Again Assumption~\ref{ass:bounded_kernel} implies $|u_{jj'}|\le 2K$. Thus this is the centered sample mean of a bounded finite population of pair values under simple random sampling of units. Equivalently, it is a finite-population U-statistic of order two with bounded kernel. By the exact variance decomposition in Theorem~\ref{thm:me_variance_compact}, its design variance is $O(m^{-1})$, because the leading projection term is $O(m^{-1})$ and the degenerate remainder is $O(m^{-2})$. Therefore
\[
\frac{\widehat A_{qq}-A_{qq}^\star}{N(N-1)}
=
O_{\mathbb P_L}(m^{-1/2}).
\]
This proves Equation~\eqref{eq:me_normalized_rates}.

We now turn to asymptotic normality. The first component in Equation~\eqref{eq:pq_ht_error} is a centered sample mean under simple random sampling without replacement, so any standard finite-population central limit theorem for bounded populations applies. For the second component, the Hoeffding decomposition used in Theorem~\ref{thm:me_variance_compact} shows that
\[
\frac{\widehat A_{qq}-A_{qq}^\star}{N(N-1)}
=
\frac{2}{m}\sum_{j\in L}\alpha_j + R_{m,N},
\]
where $\{\alpha_j\}$ is the first-order projection with bounded second moments and
\[
\Var_L(R_{m,N})=O(m^{-2}).
\]
Hence
\[
\sqrt m\,R_{m,N}\to 0
\qquad\text{in } \mathbb P_L\text{-probability},
\]
so the asymptotic distribution of the $qq$ term is governed by its projected sample-mean part. Therefore both normalized errors reduce asymptotically to linear statistics of the audit sample, and by the multivariate finite-population CLT for simple random sampling, the vector
\[
\sqrt m
\begin{pmatrix}
(\widehat A_{pq}-A_{pq}^\star)/(nN)\\[3pt]
(\widehat A_{qq}-A_{qq}^\star)/(N(N-1))
\end{pmatrix}
\]
is asymptotically Gaussian whenever its finite-population covariance matrix converges to a finite limit.

Finally,
\[
\widehat{\mathrm{MMD}}^2_{\mathrm{MEC}}
-
\widehat{\mathrm{MMD}}_\star^2
=
\frac{\widehat A_{qq}-A_{qq}^\star}{N(N-1)}
-
2\,\frac{\widehat A_{pq}-A_{pq}^\star}{nN}.
\]
Thus Equation~\eqref{eq:me_oracle_clt} follows immediately from the joint Gaussian convergence and the continuous mapping theorem, with $V_{\mathrm{design}}$ equal to the asymptotic variance of this linear combination.
\end{proof}

\begin{proof}[Proof of Theorem~\ref{thm:me_consistency_population}.]
Decompose
\begin{equation}
\label{eq:population_consistency_decomp}
\widehat{\mathrm{MMD}}^2_{\mathrm{MEC}} - \mathrm{MMD}^2(p,q;k) =
\underbrace{
\left(\widehat{\mathrm{MMD}}^2_{\mathrm{MEC}} - \widehat{\mathrm{MMD}}_\star^2 \right)
}_{(\mathrm I)} + 
\underbrace{
\left( \widehat{\mathrm{MMD}}_\star^2 - \mathrm{MMD}^2(p,q;k) \right)
}_{(\mathrm{II})}.
\end{equation}

We first control term $(\mathrm I)$. By Theorem~\ref{thm:me_design_clt},
\[
\widehat{\mathrm{MMD}}^2_{\mathrm{MEC}} - \widehat{\mathrm{MMD}}_\star^2 = O_{\mathbb P_L}(m^{-1/2})
\]
conditional on the realized finite population
$\{(Y_j,\widetilde Y_j)\}_{j=1}^N$ and $\{X_i\}_{i=1}^n$. Since the conditional
$O_{\mathbb P_L}(m^{-1/2})$ bound is uniform under Assumption~\ref{ass:bounded_kernel},
it follows by iterated expectation that
\begin{equation}
\label{eq:I_rate}
(\mathrm I)=O_{\mathbb P}(m^{-1/2}).
\end{equation}

Now consider term $(\mathrm{II})$. By Assumption~\ref{ass:additive_me_model},
the oracle statistic $\widehat{\mathrm{MMD}}_\star^2$ is exactly the standard unbiased empirical MMD estimator computed from the clean samples
\[
X_1,\dots,X_n \overset{iid}{\sim} p,
\qquad
Y_1,\dots,Y_N \overset{iid}{\sim} q.
\]
Write
\[
\widehat{\mathrm{MMD}}_\star^2 = \frac{A_{pp}^\star}{n(n-1)} + \frac{A_{qq}^\star}{N(N-1)} - 2\,\frac{A_{pq}^\star}{nN}.
\]
Under Assumption~\ref{ass:bounded_kernel}, each kernel term is bounded.

First, $A_{pp}^\star/[n(n-1)]$ is a bounded U-statistic of order two based on
$\{X_i\}_{i=1}^n$, so
\[
\frac{A_{pp}^\star}{n(n-1)}-\mathbb E[k(X,X')] = O_{\mathbb P}(n^{-1/2}).
\]

Second, $A_{qq}^\star/[N(N-1)]$ is a bounded U-statistic of order two based on
$\{Y_j\}_{j=1}^N$, so
\[
\frac{A_{qq}^\star}{N(N-1)}-\mathbb E[k(Y,Y')] = O_{\mathbb P}(N^{-1/2}).
\]

Third, $A_{pq}^\star/(nN)$ is a bounded two-sample V-statistic average of iid terms
$k(X_i,Y_j)$, hence
\[
\frac{A_{pq}^\star}{nN}-\mathbb E[k(X,Y)] = O_{\mathbb P}(n^{-1/2}+N^{-1/2}).
\]
Combining these three displays yields
\begin{equation}
\label{eq:II_rate}
(\mathrm{II}) = O_{\mathbb P}(n^{-1/2}+N^{-1/2}).
\end{equation}
In particular, $(\mathrm{II})\xrightarrow{\mathbb P}0$, which is the usual consistency of the clean-sample empirical MMD (see \citep[][]{gretton2012kernel}).

Finally, combining \eqref{eq:I_rate}, \eqref{eq:II_rate}, and
\eqref{eq:population_consistency_decomp},
\[
\widehat{\mathrm{MMD}}^2_{\mathrm{MEC}} - \mathrm{MMD}^2(p,q;k) = O_{\mathbb P}(n^{-1/2}+N^{-1/2}+m^{-1/2}),
\]
which proves \eqref{eq:me_population_rate}. Since
$n/N\to\gamma\in(0,\infty)$ and $m/N\to\rho\in(0,1]$ by
Assumption~\ref{ass:asymptotic_regime}, the three rates are all of order
$N^{-1/2}$, and therefore
\[
\widehat{\mathrm{MMD}}^2_{\mathrm{MEC}} - \mathrm{MMD}^2(p,q;k) = O_{\mathbb P}(N^{-1/2}).
\]
This implies \eqref{eq:me_population_consistency}.
\end{proof}

\section{Non-uniform and Biased Audit Sampling}
\label{sec:nonuniform-audit}

The present analysis assumes that the audit set is drawn uniformly at random, independently of the realized data. Under this simple random audit design,
\[
\Pr(j\in L)=\frac{m}{N},
\qquad
\Pr(j,j'\in L)=\frac{m(m-1)}{N(N-1)},
\]
for all distinct $j\neq j'$. This assumption is useful because it leads to a particularly simple Horvitz--Thompson correction: each audited unit receives the same weight $N/m$, and each audited pair receives the same weight $N(N-1)/\{m(m-1)\}$.

In many scientific workflows, however, audits are not collected uniformly. Validation measurements may be targeted toward observations with high uncertainty, large residuals, unusual covariates, high scientific value, or greater ease of manual verification. Such designs can be statistically efficient, since they concentrate audit effort on informative observations, but they can also introduce selection bias if the audit mechanism is ignored. In particular, the empirical distribution of audited units may no longer represent the full target sample. Consequently, applying the uniform-audit MEC-MMD correction to a non-uniform audit sample may over-correct or under-correct the kernel sums.

A natural extension is to replace the uniform correction factors by inverse-probability weights. Let
\[
\pi_j=\Pr(j\in L),
\qquad
\pi_{jj'}=\Pr(j,j'\in L)
\]
denote the first- and second-order inclusion probabilities under the audit design. We assume that these probabilities are known, or can be consistently estimated, and satisfy the positivity conditions
\[
\pi_j>0,
\qquad
\pi_{jj'}>0
\]
for all units and all distinct pairs whose corrections may be needed. Then the clean--target kernel sum can be corrected by
\[
\widehat A^{\mathrm{IPW}}_{pq}
=
\sum_{i=1}^n\sum_{j=1}^N k(X_i,\widetilde Y_j)
+
\sum_{i=1}^n\sum_{j\in L}
\frac{1}{\pi_j}
\left\{
k(X_i,Y_j)-k(X_i,\widetilde Y_j)
\right\}.
\]
Similarly, the target--target kernel sum can be corrected by
\[
\widehat A^{\mathrm{IPW}}_{qq}
=
\sum_{j\neq j'} k(\widetilde Y_j,\widetilde Y_{j'})
+
\sum_{\substack{j,j'\in L\\ j\neq j'}}
\frac{1}{\pi_{jj'}}
\left\{
k(Y_j,Y_{j'})-k(\widetilde Y_j,\widetilde Y_{j'})
\right\}.
\]
The resulting IPW version of MEC-MMD is
\[
\widehat{\mathrm{MMD}}^{2}_{\mathrm{IPW}}
=
\frac{A^\star_{pp}}{n(n-1)}
+
\frac{\widehat A^{\mathrm{IPW}}_{qq}}{N(N-1)}
-
2\frac{\widehat A^{\mathrm{IPW}}_{pq}}{nN}.
\]

The design-unbiasedness argument is the same as in the uniform case. Conditional on the realized finite population,
\[
\mathbb E_L
\left[
\sum_{j\in L}
\frac{1}{\pi_j}
\left\{
k(X_i,Y_j)-k(X_i,\widetilde Y_j)
\right\}
\right]
=
\sum_{j=1}^N
\left\{
k(X_i,Y_j)-k(X_i,\widetilde Y_j)
\right\},
\]
and
\[
\mathbb E_L
\left[
\sum_{\substack{j,j'\in L\\ j\neq j'}}
\frac{1}{\pi_{jj'}}
\left\{
k(Y_j,Y_{j'})-k(\widetilde Y_j,\widetilde Y_{j'})
\right\}
\right]
=
\sum_{j\neq j'}
\left\{
k(Y_j,Y_{j'})-k(\widetilde Y_j,\widetilde Y_{j'})
\right\}.
\]
Therefore,
\[
\mathbb E_L\!\left[
\widehat A^{\mathrm{IPW}}_{pq}
\right]
=
A^\star_{pq},
\qquad
\mathbb E_L\!\left[
\widehat A^{\mathrm{IPW}}_{qq}
\right]
=
A^\star_{qq},
\]
and hence
\[
\mathbb E_L
\left[
\widehat{\mathrm{MMD}}^{2}_{\mathrm{IPW}}
\right]
=
\widehat{\mathrm{MMD}}^{2}_{\star}.
\]
Thus, the simple-random-audit estimator studied in this paper is a special case of an IPW correction with
\[
\pi_j=\frac{m}{N},
\qquad
\pi_{jj'}=\frac{m(m-1)}{N(N-1)}.
\]
\paragraph{Self-normalised (H\'ajek) version.}
Let $w_j=1/\pi_j$ and, for independent (Poisson) inclusion, $1/\pi_{jj'}=w_jw_{j'}$. Define $S_1=\sum_{j\in L}w_j$ and $S_2=\sum_{j,j'\in L,\,j\neq j'}w_jw_{j'}=S_1^2-\sum_{j\in L}w_j^2$. The H\'ajek version replaces the normalisers $N$ and $N(N-1)$ of the audit corrections by $S_1$ and $S_2$:
\begin{align*}
\widehat{\mathrm{MMD}}^{2}_{\mathrm{IPW\text{-}H}}
&=\frac{A^\star_{pp}}{n(n-1)}
+\frac{1}{N(N-1)}\sum_{j\neq j'}k(\widetilde Y_j,\widetilde Y_{j'})
+\frac{1}{S_2}\sum_{\substack{j,j'\in L\\ j\neq j'}}w_jw_{j'}\,\Gamma_{jj'}\\
&\quad-\frac{2}{nN}\sum_{i=1}^n\sum_{j=1}^N k(X_i,\widetilde Y_j)
-\frac{2}{n S_1}\sum_{i=1}^n\sum_{j\in L}w_j\,\Lambda_{ij}.
\end{align*}
With equal weights it equals $\widehat{\mathrm{MMD}}^2_{\mathrm{MEC}}$. It is a ratio estimator, so it is not exactly unbiased in finite samples (exact unbiasedness is for the plain IPW version above), and we do not analyse it theoretically.
This extension would broaden MEC-MMD to two-phase sampling, stratified validation, active auditing, and outcome-dependent audit designs. For example, one could oversample units with large estimated measurement uncertainty, rare covariate values, or high leverage in the MMD objective, while still preserving design-unbiasedness by weighting each audited correction by its inverse inclusion probability. Such designs may be more efficient than uniform auditing when audit resources are limited.

The main challenge is variance. If some inclusion probabilities are small, the corresponding inverse-probability weights can be large, leading to unstable corrections. This issue is especially important for the $qq$ term, since it depends on second-order inclusion probabilities $\pi_{jj'}$. Another open question is how to estimate the inclusion probabilities when the audit design is not fully known, for example, when validation data are collected opportunistically or through expert-driven selection.

Developing stable IPW, stabilized-weight, and doubly robust versions of MEC-MMD under non-uniform audit sampling is therefore an important direction for future work. Such extensions would preserve the central idea of MEC-MMD, i.e., correcting MMD at the kernel level using audited latent values, while allowing the audit mechanism to reflect realistic scientific data-collection workflows.

\section{Detail of Experiments} \label{app:details_experiments}

\paragraph{Computational cost.}
Let $n$ be the number of generated samples, $N$ the number of noisy observations and $m$ the audit size. The plug-in terms of $\widehat{\mathrm{MMD}}^2_{\mathrm{MEC}}$ need $O(n^2+N^2+nN)$ kernel evaluations, the same as the naive estimator. The audit correction adds $m(m-1)/2$ evaluations of $\Gamma_{jj'}$ (two kernel evaluations each) and $2nm$ evaluations for the $\Lambda_{ij}$, i.e.\ $O(m^2+nm)$ with $m\ll N$. Using a minibatch of $B$ noisy observations per step with the full audit correction is unbiased for the same objective and costs $O(n^2+nB+nm)$.

\textit{Benchmark.} We time one evaluation of the objective and its gradient with respect to $n=1000$ generated samples in $d=2$ dimensions (Gaussian kernel, $m=0.1N$, float32) in a standalone PyTorch implementation on Mac M1 Pro, reporting the median and range of 4 repeated runs of 12 evaluations each after warm-up. ``Recompute'' evaluates every term at each step (naive MMD and MEC-MMD), ``cached'' precomputes the terms that do not involve the generated samples and ``minibatch'' additionally uses $B=2048$ noisy observations per step. The model's own forward and backward cost is not included and it is identical for naive MMD and MEC-MMD.

\textit{Results.} Table~\ref{tab:runtime} shows that precomputing the terms that do not involve the generated sample makes a MEC-MMD step about $2.7\times$ faster at $N=5000$ and about $5\times$ faster at $N=10^4$, and that a minibatch of $B=2048$ observations is faster by a further factor of about $1.5$--$2$. The extra work of the audit correction ($O(m^2+nm)$ kernel evaluations) is not distinguishable from run-to-run variation. The time of a MEC-MMD step differed from that of naive MMD by $+7\%$ to $+13\%$ at $N=5000$ and by $-28\%$ to $+24\%$ at $N=10^4$ when both recompute every term.
\begin{table}[h]
\centering
\caption{Time per evaluation of the objective and its gradient (ms; median over 4 runs, with the range below) on one thread of an Apple M1 Pro CPU ($n=1000$ generated samples, $d=2$, $m=0.1N$, float32). Recompute: every term at each step; cached: model-independent terms precomputed; minibatch: $B=2048$ noisy observations per step.}
\label{tab:runtime}
\small
\setlength{\tabcolsep}{5pt}
\begin{tabular}{rrcccc}
\toprule
 & & \multicolumn{2}{c}{Recompute} & Cached & Minibatch \\
\cmidrule(lr){3-4}\cmidrule(lr){5-5}\cmidrule(lr){6-6}
$N$ & $m$ & Na\"ive & MEC-MMD & MEC-MMD & MEC-MMD \\
\midrule
5000 & 500 & 282 & 304 & 113 & 76 \\
 & & {\scriptsize [259--294]} & {\scriptsize [289--320]} & {\scriptsize [101--159]} & {\scriptsize [55--106]} \\[3pt]
10000 & 1000 & 943 & 992 & 190 & 89 \\
 & & {\scriptsize [855--1373]} & {\scriptsize [940--1057]} & {\scriptsize [152--285]} & {\scriptsize [77--115]} \\
\bottomrule
\end{tabular}
\end{table}
\paragraph{Numerical check of the design-based results.}
\textit{Setting.}  $X_i\sim\mathcal N(3,1)$ ($n=500$), $Y_j\sim\mathcal N(0,1)$ ($N=500$), $\widetilde Y_j=Y_j+\varepsilon_j$ with $\varepsilon_j\sim\mathcal N(0.1,0.4^2)$, and the Gaussian kernel $k(a,b)=\exp\{-(a-b)^2/(2h^2)\}$ with $h=1$. We generate 40 independent populations. For each population we compute the oracle estimate $\widehat{\mathrm{MMD}}^2_\star$ and the variance $\mathrm{Var}_{\mathrm{theo}}$ of Theorem~\ref{thm:me_variance_compact} from the realised $\Gamma_{jj'}$ and $\Lambda_{ij}$, and then draw 3000 simple random audits of each size $m\in\{5,20,150\}$.

\textit{Statistics.} For each audit we compute $\widehat{\mathrm{MMD}}^2_{\mathrm{MEC}}$ and $Z=(\widehat{\mathrm{MMD}}^2_{\mathrm{MEC}}-\widehat{\mathrm{MMD}}^2_\star)/\sqrt{\mathrm{Var}_{\mathrm{theo}}}$. Table~\ref{tab:clt_check} reports, averaged over the 40 populations, the mean of $\widehat{\mathrm{MMD}}^2_{\mathrm{MEC}}$ over audits, its empirical variance over audits, $\mathrm{Var}_{\mathrm{theo}}$, the standard deviation and skewness of $Z$, and the fraction of audits with $|Z|\le1.96$.

\textit{Estimated variance.} The last column replaces $\mathrm{Var}_{\mathrm{theo}}$ by $\widehat{\mathrm{Var}}=\frac{4(N-m)}{mN}\widehat S^2_{\mathrm{proj}}$, the leading term of Theorem~\ref{thm:me_variance_compact}, with $\widehat S^2_{\mathrm{proj}}=\frac1{m-1}\sum_{j\in L}\hat u_j^2$ and $\hat u_j=\frac{N-1}{N-2}(\hat\Gamma_j-\hat\Gamma)-(\Lambda_j-\bar\Lambda_L)$, where $\hat\Gamma_j=\frac1{m-1}\sum_{j'\in L\setminus\{j\}}\Gamma_{jj'}$, $\hat\Gamma=\frac1m\sum_{j\in L}\hat\Gamma_j$ and $\bar\Lambda_L=\frac1m\sum_{j\in L}\Lambda_j$. It uses only audited quantities and ignores the degenerate term, which is of order $m^{-2}$.

\textit{Results.} In this setting the mean of $\widehat{\mathrm{MMD}}^2_{\mathrm{MEC}}$ matches the oracle at every audit size (Theorem~\ref{thm:me_unbiased_sums}), the empirical variance matches $\mathrm{Var}_{\mathrm{theo}}$ (Theorem~\ref{thm:me_variance_compact}), and $Z$ has unit standard deviation, negligible skewness and 95\% coverage already at $m=5$ (Theorem~\ref{thm:me_design_clt}). These are the true-variance quantities of the theorems. With the variance estimated from the audit, coverage is [0.84, 0.92, 0.95] at $m=5,20,150$, so very small audits give less reliable intervals.

\begin{table}[h]
\centering
\caption{Numerical check of Theorems~\ref{thm:me_unbiased_sums}, \ref{thm:me_variance_compact} and \ref{thm:me_design_clt}; 40 populations, 3000 simple random audits each, mean over populations. $Z=(\widehat{\mathrm{MMD}}^2_{\mathrm{MEC}}-\widehat{\mathrm{MMD}}^2_\star)/\sqrt{\mathrm{Var}_{\mathrm{theo}}}$, with $\mathrm{Var}_{\mathrm{theo}}$ the variance of Theorem~\ref{thm:me_variance_compact} on the realised population; the last column standardises by a variance estimated from the audit only. Oracle $\widehat{\mathrm{MMD}}^2_\star=0.894$.}
\label{tab:clt_check}
\begin{tabular}{rccccccc}
\toprule
$m$ & $\mathbb E[\widehat{\mathrm{MMD}}^2_{\mathrm{MEC}}]$ & $\mathrm{Var}_{\mathrm{emp}}$ & $\mathrm{Var}_{\mathrm{theo}}$ & $\mathrm{sd}(Z)$ & skew & 95\% cov. & 95\% cov.\ (est.\ var.) \\
\midrule
5 & 0.8937 & $1.62\times10^{-2}$ & $1.63\times10^{-2}$ & 1.00 & 0.01 & 0.948 & 0.836 \\
20 & 0.8941 & $3.29\times10^{-3}$ & $3.27\times10^{-3}$ & 1.00 & 0.00 & 0.949 & 0.924 \\
150 & 0.8942 & $3.02\times10^{-4}$ & $3.03\times10^{-4}$ & 1.00 & $-0.01$ & 0.950 & 0.947 \\
\bottomrule
\end{tabular}
\end{table}

\paragraph{Experiment 1.} This experiment is designed to explicitly separate bias and variance in the additive measurement-error setting. We generate a clean reference sample
\[
X_1,\dots,X_n \overset{\mathrm{iid}}{\sim} \mathcal N(3,1)
\]
and a latent target sample
\[
Y_1,\dots,Y_N \overset{\mathrm{iid}}{\sim} \mathcal N(0,1),
\]
with $n=N=500$. The target quantity is the squared population discrepancy $\mathrm{MMD}^2(p,q;k)$, where $k$ is a Gaussian kernel with bandwidth $0.1$. The latent target sample is not directly available to the analyst. Instead, for each $j$,
\[
\widetilde Y_j = Y_j + \varepsilon_j,
\qquad
\varepsilon_j \overset{\mathrm{iid}}{\sim} \mathcal N(0.1,0.4^2),
\]
so the observed target sample is systematically shifted and corrupted by additive noise. We additionally draw an audit subset $L \subset \{1,\dots,N\}$ uniformly without replacement, with
\[
m \in \{20,40,60,80,100,125,150\},
\]
and reveal the latent values $\{Y_j : j \in L\}$ for those audited points.

For each audit size $m$, we compare four estimators. The \emph{oracle} estimator computes empirical MMD using the full latent sample $Y$, and is unavailable in practice but provides a benchmark corresponding to the smallest variance achievable when all target values are observed cleanly. The \emph{naive noisy} estimator computes empirical MMD directly from $(X,\widetilde Y)$, ignoring measurement error entirely. Because this estimator uses all $N$ target observations, it can have relatively small variance, but it is biased for $\mathrm{MMD}^2(p,q;k)$ since it targets the distribution of the noisy proxies rather than the latent distribution $q$. The \emph{audit-only} estimator uses only the audited latent target points together with the full clean sample $X$. This removes the measurement-error bias, but its effective target sample size is only $m$, so its variance is substantially larger than that of the oracle benchmark. Finally, the proposed MEC-MMD estimator combines the full noisy sample with an audit-based Horvitz--Thompson correction. Its role is precisely to exploit the low-variance information in the noisy sample while using the audit sample only to remove the induced bias.

We repeat the experiment over $5000$ independent Monte Carlo realizations. For each method and each audit size, we compute the empirical bias and variance of the resulting estimator. The appendix figure reports these two components in separate panels. This decomposition makes the tradeoff transparent. The naive noisy estimator occupies the low-variance but biased extreme. The audit-only estimator is essentially unbiased, but its variance is much larger because it discards almost all target observations. MEC-MMD improves on both in the expected direction: compared with the naive estimator, it eliminates most or all of the measurement-error bias; compared with the audit-only estimator, it achieves substantially lower variance by leveraging the full noisy sample. Across all audit sizes, the oracle curve provides the natural lower benchmark for variance, while the gap between the naive and corrected methods in the bias panel visualizes the cost of ignoring measurement error.

\begin{figure}[th!]
    \centering
    \includegraphics[width=0.98\linewidth]{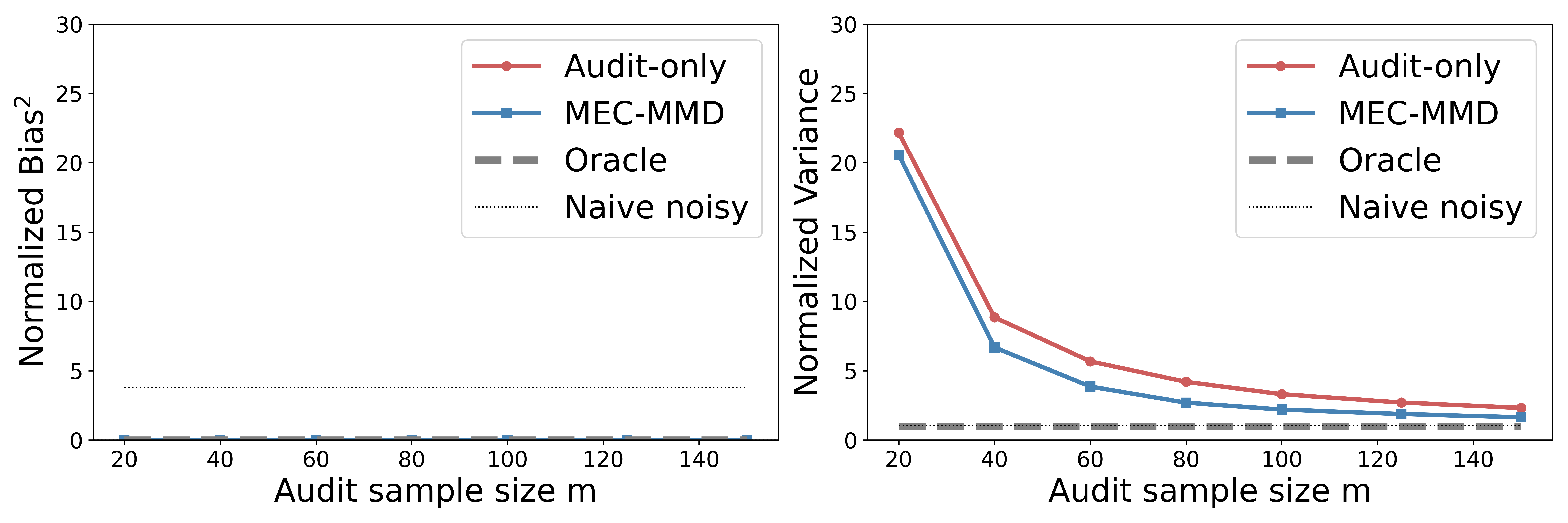}
    \caption{Bias (left panel) and variance (right panel) for the Gaussian measurement-error experiment in Figure~\ref{fig:me_main_result}. The left panel reports empirical bias relative to $\mathrm{MMD}^2(p,q;k)$, and the right panel reports empirical variance. The na\"ive noisy estimator has the smallest variance among practical methods because it uses all $N$ observed target points, but it is biased because it ignores measurement error. The audit-only estimator is unbiased but has substantially larger variance since its effective target sample size is only $m$. MEC-MMD uses the noisy sample to retain efficiency and the audit sample to remove bias, yielding a markedly better bias--variance tradeoff: much lower variance than audit-only estimation and far smaller bias than the naive noisy estimator. The oracle variance serves as the lower benchmark corresponding to full access to the latent target sample.}
    \label{fig:me_bias_variance_appendix}
\end{figure}

\paragraph{Experiment 2 -- Effect of measurement noise.}
To study the effect of measurement noise on estimator efficiency, we consider a one-dimensional Gaussian benchmark. We generate a clean reference sample
\[
X_1,\dots,X_n \overset{\mathrm{iid}}{\sim} \mathcal N(3,1),
\]
and a latent target sample
\[
Y_1,\dots,Y_N \overset{\mathrm{iid}}{\sim} \mathcal N(0,1),
\]
with $n=N=500$. The estimand is the squared population discrepancy $\mathrm{MMD}^2(p,q;k)$, computed using a Gaussian kernel with bandwidth $0.1$.

The analyst does not observe the latent targets $Y_j$ directly. Instead, noisy surrogates are observed,
\[
\widetilde Y_j = Y_j + \varepsilon_j,
\qquad
\varepsilon_j \sim \mathcal N(\mu_\varepsilon, \sigma_\varepsilon^2).
\]
Throughout the experiment we fix $\mu_\varepsilon = 0.1$ and vary the noise
standard deviation across $\sigma_\varepsilon \in \{0.0,\,0.1,\,0.2,\,0.3,\,0.4,\,0.5\}$. An audit subset $L \subset \{1,\dots,N\}$ is drawn uniformly without
replacement with sizes $m \in \{20,40,60,80,100,125,150\}$. For audited points, the latent values $Y_j$ are revealed. The experiment is repeated over $5000$ independent Monte Carlo realizations. For each configuration, we evaluate three practical estimators together with an oracle benchmark:

\begin{itemize}
\item \textbf{Oracle MMD:} empirical MMD computed using the full latent sample $Y_1,\dots,Y_N$. This estimator is not available in practice but provides the minimum achievable variance when all targets are observed cleanly.

\item \textbf{Audit-only:} empirical MMD computed using the clean reference sample $X$ and only the audited latent targets $\{Y_j : j \in L\}$. This estimator is unbiased but its effective target sample size is $m$, leading to substantially higher variance when $m \ll N$.

\item \textbf{MEC-MMD:} the proposed measurement-error-corrected estimator, which uses the full noisy sample $\widetilde Y$ together with audit-based Horvitz--Thompson corrections to recover unbiased kernel sums.
\end{itemize}

For each estimator we compute empirical mean, bias, and variance across the $5000$ realizations.  

The results illustrate how measurement noise affects the efficiency of the correction procedure. As the noise level increases, the variance of MEC-MMD increases because the correction terms become larger and more variable. However, even at the highest noise level considered, MEC-MMD remains substantially more efficient than the audit-only estimator. At the same time, when the noise level approaches zero, MEC-MMD becomes nearly identical to the oracle estimator since the observed proxies coincide with the latent targets. This experiment therefore highlights the key advantage of MEC-MMD: it leverages the large noisy dataset for efficiency while using the audit sample only to remove the bias induced by measurement error.

\paragraph{Experiment 3 -- Parameter estimation under measurement error.}
We consider a two-dimensional generative model in which latent target samples lie on a noisy circle with unknown radius $r_\star$. The true latent points are generated as
\[
\theta_i \sim \mathrm{Unif}(0,2\pi), \qquad
R_i = r_\star + \eta_i, \qquad
\eta_i \sim \mathcal N(0,\sigma_r^2),
\]
\[
Y_i = (R_i \cos\theta_i,\, R_i \sin\theta_i),
\]
with $r_\star = 2.0$ and radial noise standard deviation $\sigma_r = 1.0$. This produces a circular ring whose thickness reflects variability in the underlying radius.

The analyst does not observe the latent targets $Y_i$ directly. Instead, noisy proxies are observed,
\[
\widetilde Y_i = Y_i + \varepsilon_i,
\qquad
\varepsilon_i \sim \mathcal N(0,\sigma_\varepsilon^2 I_2),
\]
with observation noise standard deviation $\sigma_\varepsilon = 1.0$. The observed target sample size is $N=1000$. Additionally, a small audit subset of size $m=50$ is drawn uniformly without replacement, revealing the latent values for those points.

\textit{Model estimation procedure.} To estimate the unknown radius, we consider a grid of candidate radii $r \in [0,4]$, discretized using $400$ evenly spaced values. For each candidate radius, we generate a reference sample of size $n_\text{model}=20{,}000$ from the same noisy-circle model but with radius fixed at $r$. We then compute the empirical MMD between the reference sample and the target data using a Gaussian kernel with bandwidth $2.0$. Four estimators are compared:
\begin{itemize}
\item \textbf{Na\"ive MMD:} empirical MMD between the candidate model sample and the noisy observations $\widetilde Y$.
\item \textbf{Audit-only MMD:} empirical MMD between the candidate model sample and the audited latent targets.
\item \textbf{Oracle MMD:} empirical MMD between the candidate model sample and the full latent sample $Y$.
\item \textbf{MEC-MMD:} the proposed measurement-error-corrected estimator that combines the noisy observations with Horvitz--Thompson corrections computed from the audit sample.
\end{itemize}
For each estimator, the estimated radius is obtained by minimizing the corresponding empirical MMD curve over the candidate grid.

The right panel of Figure~\ref{fig:me_main_result} shows the MMD objective as a function of the candidate radius. The oracle curve reaches its minimum near the true radius $r_\star = 2.0$. The na\"ive estimator produces a shifted minimum due to measurement error in the observations, demonstrating the bias induced by ignoring noise. The audit-only estimator reduces this bias but produces a noisier objective because it relies on only $m=50$ samples. The MEC-MMD estimator combines the information from the noisy dataset and the audit subset, producing an objective curve that closely matches the oracle curve and correctly identifies the true radius. This experiment highlights how measurement-error correction can enable reliable parameter estimation even when the majority of observations are corrupted.

\begin{figure}
    \centering
    \includegraphics[width=0.32\linewidth]{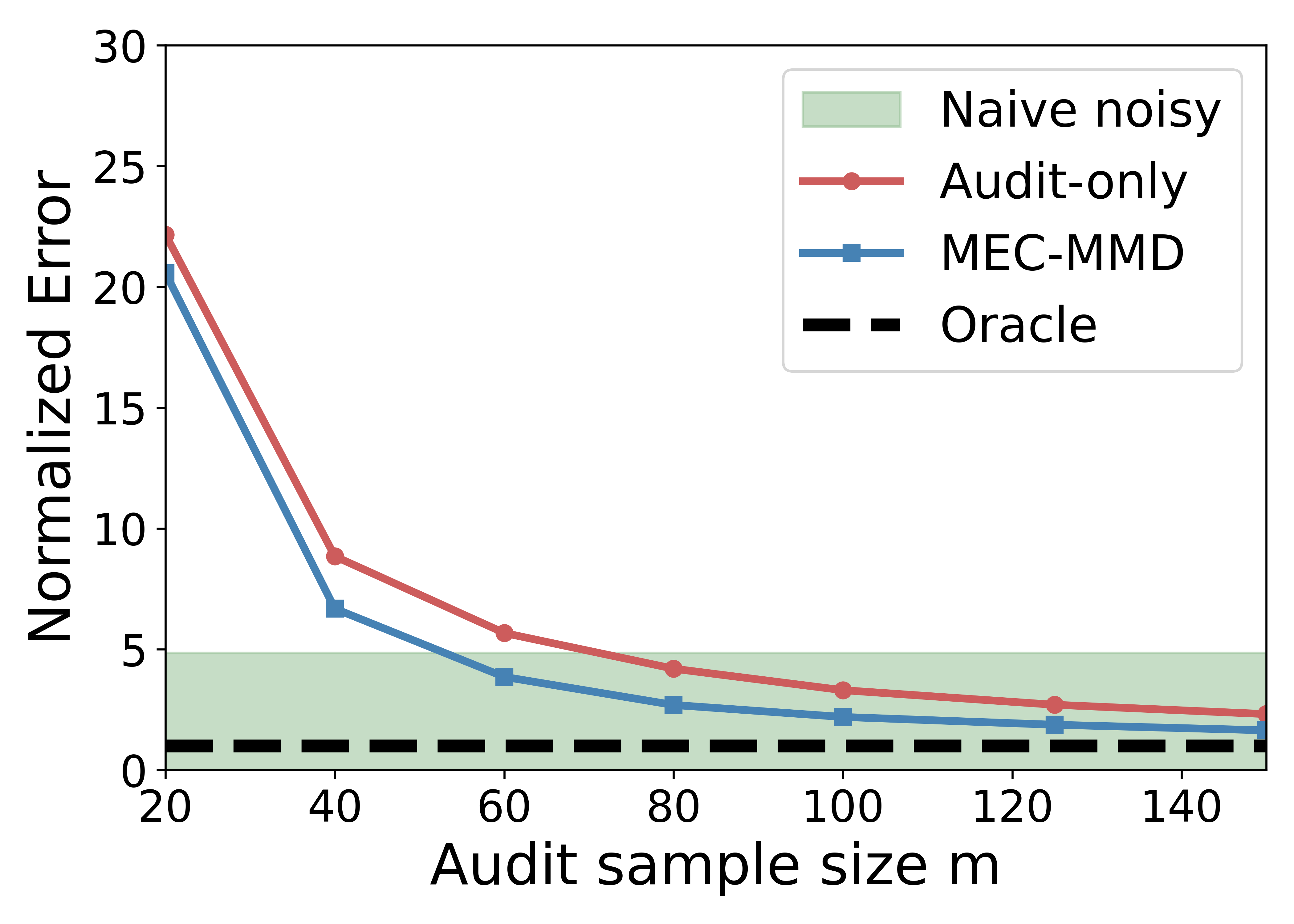}
    \includegraphics[width=0.32\linewidth]{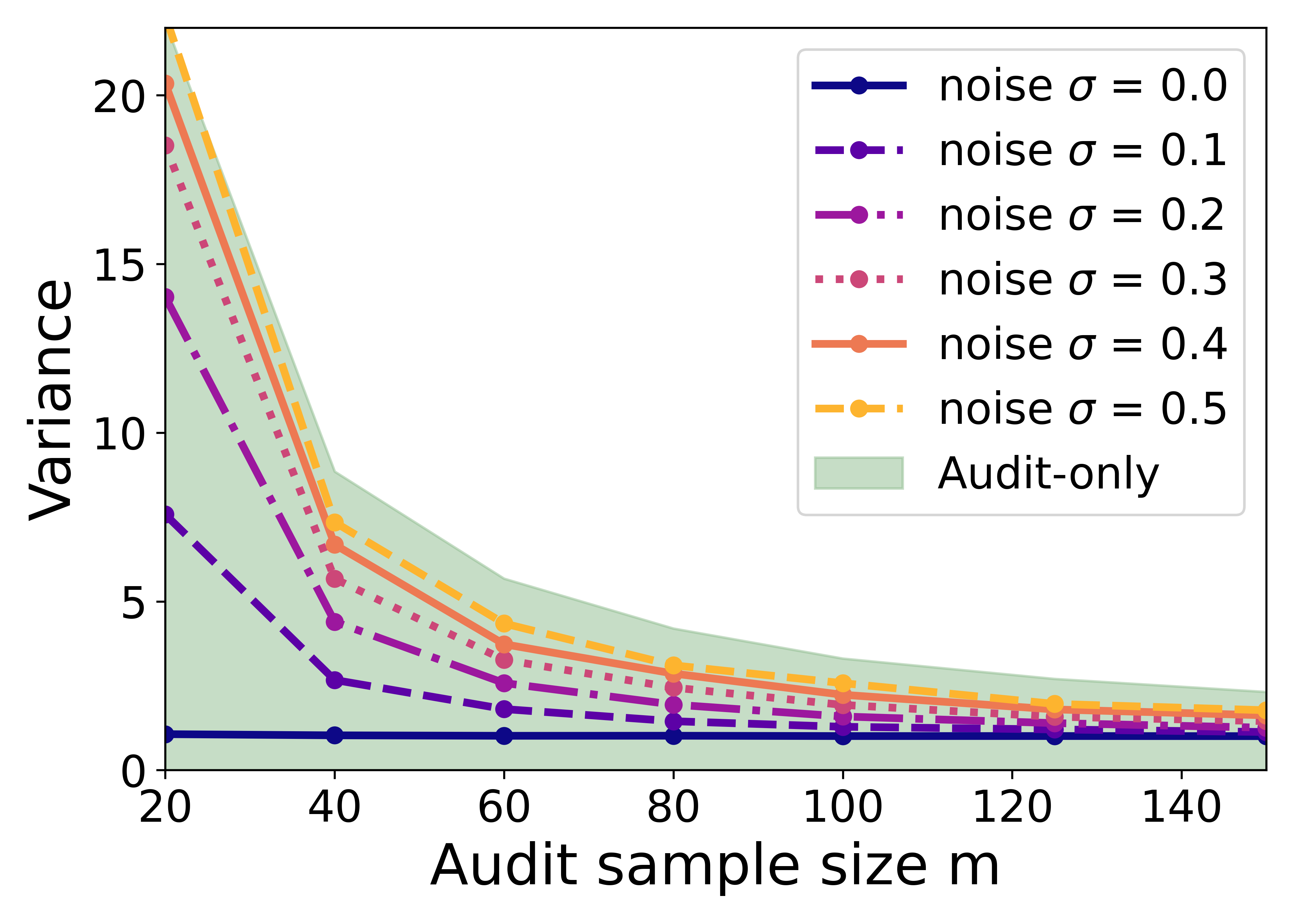}
    \includegraphics[width=0.32\linewidth]{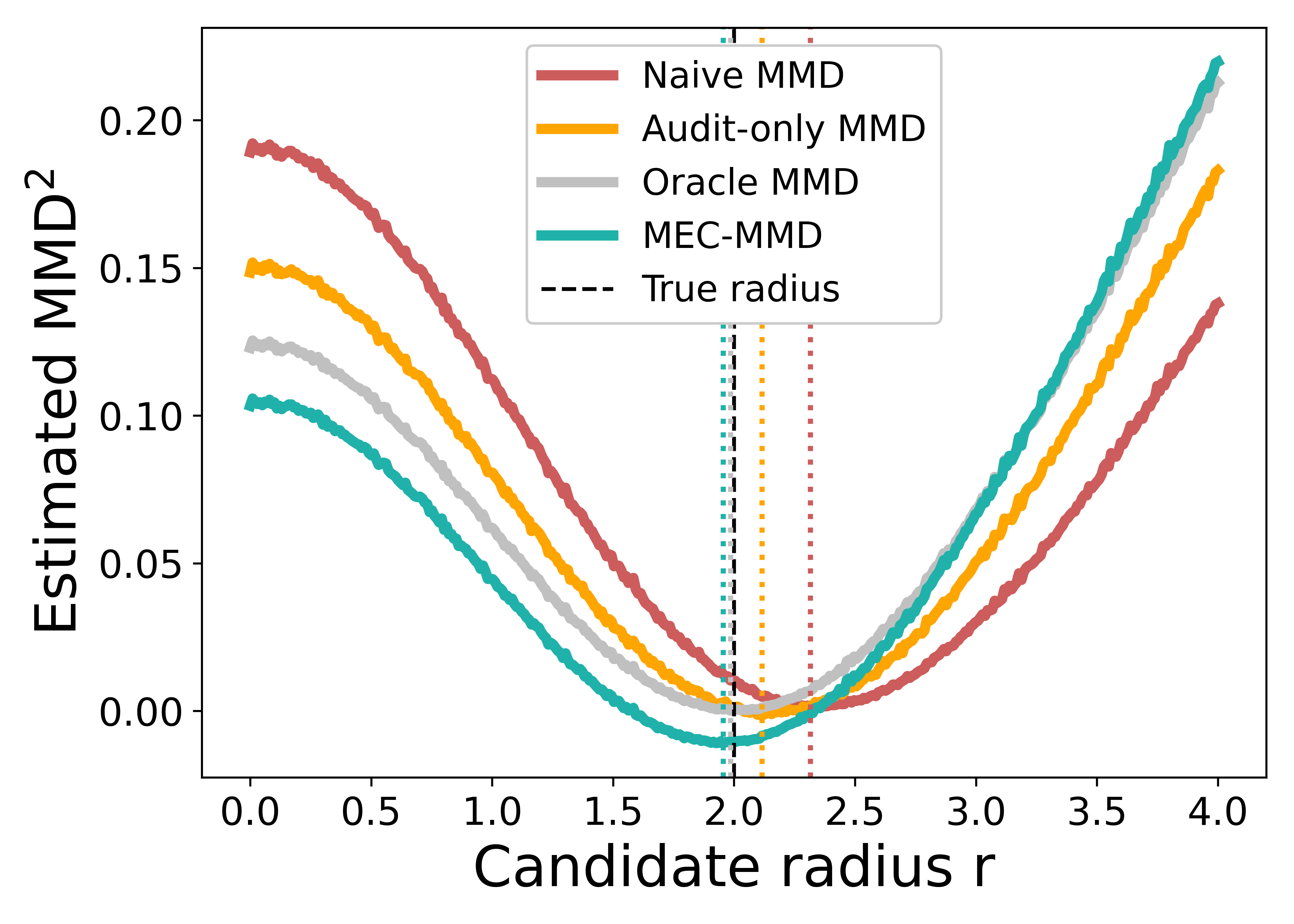}
    \caption{{\bf Left Panel.} Estimation error. The oracle estimator uses the full latent sample and yields a lower-variance bound. The naive estimator based on the full noisy sample has low variance but is biased for $\mathrm{MMD}^2(p,q;k)$. The audit-only estimator removes this bias but has a much larger variance because it uses only $m$ latent target points. MEC-MMD combines noisy samples with audit-based correction, thereby removing bias induced by noisy measurements while achieving substantially lower variance than audit-only estimation. {\bf Middle Panel.} Measurement error effect on estimator variance. The horizontal axis shows the audit sample size $m$, and the vertical axis shows the variance of the estimator. Colored curves correspond to different noise levels $\sigma \in \{0,0.1,0.2,0.3,0.4,0.5\}$. As measurement noise increases, the variance of the MEC-MMD estimator rises substantially, especially when the audit sample is small, because the correction must be estimated from a limited number of latent target points. For every noise level, increasing $m$ reduces variance, reflecting the improved precision of the audit-based correction. When $\sigma=0$, the variance equals oracle variance and is flat in $m$, since no measurement-error correction is needed. The results illustrate that measurement noise primarily affects statistical efficiency rather than identifiability: larger noise inflates uncertainty, but this effect can be mitigated by increasing the audit sample size.
    {\bf Right Panel.} Parameter estimation. The horizontal axis shows candidate radii, and the vertical axis shows the corresponding empirical MMD objective. Vertical lines indicate the minimizing radius for each estimator. The naive estimator (red) is biased because it compares candidate models to noisy observations. The audit-only estimator (orange) removes this bias but has higher variability due to the small audit sample. The proposed MEC-MMD estimator (teal) combines the full noisy dataset with audit-based corrections and closely matches the oracle estimator (gray), correctly identifying the true radius.}
    \label{fig:me_main_result_full}
\end{figure}

\paragraph{Experiment 4 -- Density Estimation under measurement error}
We consider a two-dimensional density estimation task in which the latent target samples follow a 10-component Gaussian Mixture Model (GMM) designed to resemble a cat paw. The true latent points are generated as
$$Y_{i} \sim \sum_{k=1}^{10}\pi_k \mathcal{N}(\mu_k, \Sigma_k),$$
where the mixture components represent four elliptical toe pads on an upper arc and three lower palm pads. The mixture weights $\pi_k$ are set to approximately $0.113$ for each toe, $0.151$ for the side pads, and $0.245$ for the central pad, with corresponding geometric means and diagonal covariances.
The analyst does not observe the latent targets $Y_{i}$ directly. Instead, noisy samples are observed,
$$\widetilde{Y}_{i} = Y_{i} + \varepsilon_{i},$$
We consider three distinct heteroscedastic noise settings. For (1) \textbf{Gaussian noise}, $\varepsilon_i \sim \mathcal{N}(0, \sigma_i^2 I_2)$ with individual scale parameters sampled as $\sigma_i \sim \text{Unif}(0.2, 0.6)$. For (2) \textbf{Laplace noise}, the noise follows a zero-mean Laplace distribution with scale $b_i = \sigma_i / \sqrt{2}$, where again $\sigma_i \sim \text{Unif}(0.2, 0.6)$. For (3) \textbf{Dependent noise}, the Laplace scale is spatially dependent on the underlying latent variable, setting the noise standard deviations for each dimension $d \in \{1, 2\}$ to $\sigma_{i,d} = 0.3 |Y_{i,d}|$. The sample size is fixed at $N = 2000$. Additionally, an audit subset of size m, varying across $200$, $300$, and $400$, is drawn uniformly without replacement, revealing the true latent values for those points.
\textit{Model estimation procedure.} All MMD-based methods fit the same density model, which is a mixture of $10$ full-covariance Gaussians in $\mathbb R^2$ with trainable weights, means and Cholesky factors. Every method, including the oracle, starts from the same scikit-learn \texttt{GaussianMixture} fit (full covariance, 5 restarts) to the noisy sample $\widetilde Y$. To estimate the underlying density, we evaluate the methods based on the Integrated Square Error (ISE). For MMD-based methods, the kernel bandwidth $h$ is selected dynamically using a stabilized quantile-based heuristic over pairwise distances. For $q\in\{0.1,\dots,0.9\}$, $h_q=\sqrt{Q_q}/(\log_{10}(2n)\sqrt d)$, where $Q_q$ is the $q$-quantile of the pairwise squared distances above $10^{-5}$ among the training sample ($Y$ for the oracle, $\widetilde Y$ otherwise) pooled with $n$ draws from the initial model; $n=2000$, $d=2$.

Five estimators are compared:
\begin{itemize}
\item \textbf{Naive MMD:} empirical MMD trained directly on the noisy observations $\widetilde{Y}$ without accounting for measurement error.
\item \textbf{Oracle MMD \cite{briol2019statistical}:} empirical MMD trained directly on the full, unobserved clean latent sample $Y$.
\item \textbf{convMMD \cite{vashistha2026convolutional}:} an estimator that accounts for noise through a resampling scheme, simulating the noise distribution by bootstrapping residuals $(\widetilde{Y} - Y)$ from the audit set.
\item \textbf{XDGMM \cite{bovy2011extreme}:} Extreme Deconvolution GMM, which analytically models measurement error by assuming zero-mean Gaussian noise. We set the noise covariance to the empirical variance of the audit residuals.
\item \textbf{MEC-MMD (Ours):} the proposed measurement-error-corrected estimator utilizing the available audit samples.
\end{itemize}
Figure~\ref{fig: paw_density} and Table~\ref{tab: paw} illustrate the performance of these estimators. In Figure~\ref{fig: paw_density}, visual comparisons of the estimated density under Gaussian noise (with an audit size of $m = 400$) reveal that the naive estimator suffers from severe blurring due to the uncorrected heteroscedastic noise. Table~\ref{tab: paw} summarizes the ISE across multiple runs and noise distributions. The naive estimator and XDGMM demonstrate significant bias, with XDGMM particularly struggling under the complex, non-Gaussian noise conditions. The convMMD method offers a strong baseline but relies heavily on empirical bootstrapping. Our MEC-MMD estimator effectively combines the full noisy dataset with the audit subset, producing an estimate that closely matches the oracle performance. As the audit sample size increases to $m = 300$ and above, MEC-MMD consistently achieves the lowest ISE among the non-oracle methods across all three noise settings, highlighting its capability to accurately recover the underlying density even under complex, dependent noise.  Error bars are computed based on 10 random realizations, mean and standard deviations are reported. 

\paragraph{Experiment 4b -- Non-uniform audits.}
We repeat Experiment 4 with non-random audit samples. We vary the audit size as $m_{\mathrm{target}}\in\{200,300,400\}$.

\textit{Audit design.} Let $K=10$ be the number of mixture components and $k(j)$ the component that generated unit $j$. Each component receives an audit-intensity multiplier $c_k=\exp(u_k-\bar u)$ with $u_1,\dots,u_K\overset{\mathrm{iid}}{\sim}\mathrm{Unif}(\log 0.3,\log 3)$ and $\bar u=K^{-1}\sum_k u_k$, so the multipliers have geometric mean one. Components with $c_k<1$ are under-sampled, those with $c_k>1$ over-sampled, and the ratio of the largest to the smallest multiplier is at most $10$. The multipliers are drawn once and kept fixed across seeds. Unit $j$ is audited independently with probability
\[
\pi_j=\min\Bigl\{1,\ \max\Bigl\{10^{-3},\ \frac{m_{\mathrm{target}}\,c_{k(j)}}{\sum_{i=1}^{N}c_{k(i)}}\Bigr\}\Bigr\},
\]
so that $\pi_{jj'}=\pi_j\pi_{j'}$ for $j\neq j'$, the realised audit size $m$ is random with $\mathbb E[m]\approx m_{\mathrm{target}}$, and the inclusion probabilities are known exactly. The component label acts as an observed stratum.

\textit{Estimators.} MEC-MMD-IPW is the self-normalised (H\'ajek) estimator of Appendix~\ref{sec:nonuniform-audit} with weights $w_j=1/\pi_j$ and no weight truncation. convMMD bootstraps the audited residuals $\widetilde Y_j-Y_j$, and XDGMM uses a noise covariance with the pooled variance of the audited residuals on the diagonal; both use the audited pairs without weights. Na\"ive MMD and the oracle do not use the audit.

\textit{Protocol.} Each configuration is repeated over 10 seeds; we report the mean $\pm$ standard error of the integrated squared error on an $80\times80$ grid over $[-2.5,2.5]^2$.

\paragraph{Experiment 5 -- Denoising with Empirical Bayes in the Presence of Outliers.}

In this experiment, we demonstrate an application of MEC-MMD in empirical Bayes denoising. Given the learned density $q_{\widehat{\theta}_N}$, we use it as a data-driven prior to estimate the posterior mean $\mathbb{E}[\theta_i \mid X_i]$. We evaluate this procedure on three structured two-dimensional datasets (Two Moons, Circle, and Checkerboard), in which latent points are corrupted by heteroscedastic Gaussian noise. To assess robustness, we further contaminate the latent distribution with $3\%$ outliers. We compare our results using a normalizing-flow-based density estimation against two standard deconvolution baselines: NPEB, a nonparametric empirical Bayes method based on the NPMLE~\cite {soloff2025multivariate}, and XDGMM, which fits a Gaussian-mixture prior by expectation maximization~\citep{bovy2011extreme}.  An Oracle baseline is also included, which uses the true latent distribution estimated via KDE from a large, clean sample and thus provides an approximate upper bound on denoising performance. 

\textit{Data Generating Process and Outlier Corruption.} For each of the three topologies (Two Moons, Circle, Checkerboard), we generate $N=2000$ latent samples $\theta_i \in \mathbb{R}^2$. To evaluate robustness, we randomly select $3\%$ of the latent samples and displace them using a large uniform offset (e.g., $\mathcal{U}(-4.0, 4.0)$). This creates a set of "true" outliers that exist far outside the primary manifold.

We generate independent per-sample noise standard deviations $\sigma_i \sim \mathcal{U}(a, b)$. For the Moon and Checkerboard datasets, we use noise bounds $(a, b) = (0.2, 0.6)$. Because the Circle has a larger geometric footprint, we increase the noise bounds to $(a,b) = (0.5, 1.0)$. The final observed data is generated as $X_i = \theta_i + \varepsilon_i$, where $\varepsilon_i \sim \mathcal{N}(0, \sigma_i^2 I_2)$.

\textit{MEC-MMD.} We utilize a Masked Autoregressive Flow (MAF) as the latent model. The flow consists of 4 layers, where each layer applies a masked affine autoregressive transform parameterized by a residual network containing 2 blocks and a hidden dimension of 512, using ReLU activations. Each autoregressive layer is followed by a reverse permutation and an LU linear transformation. The base distribution is a standard bivariate Gaussian. The model is trained for 3000 epochs. We use the AdamW optimizer with a learning rate of $10^{-4}$ and a cosine decay schedule. Kernel bandwidths are chosen automatically using the median heuristic \citep{schrab2025practical,biggs2023mmd}. Once trained, we draw $M = 30,000$ samples from the learned prior $q_{\widehat{\theta}_N}$ and compute the posterior mean for each observation using importance sampling:
$$ \mathbb{E}[\theta_i \mid X_i] \approx \sum_{m=1}^M w_{i,m} \tilde{\theta}_m, \quad \text{where} \quad w_{i,m} \propto \exp\left(-\frac{\|X_i - \tilde{\theta}_m\|^2}{2\sigma_i^2}\right) $$

\textit{NPEB.} We use the Nonparametric Empirical Bayes framework via the `npeb` package. This method computes the Nonparametric Maximum Likelihood Estimator (NPMLE) using the `GLMixture` solver. The algorithm places a fine grid over the observational space and optimizes the mixture weights using convex optimization (via CVXPY). We use diagonal precision matrices parameterized by the known heteroscedastic noise variances. The denoised estimates are then extracted using the exact discrete posterior mean (GMLEB).

\textit{XDGMM.} Extreme Deconvolution (XDGMM) fits a continuous Gaussian Mixture Model to data with known heteroscedastic measurement noise. To ensure a fair comparison and avoid bottlenecks, we implemented a custom JAX-accelerated, JIT-compiled version of the XDGMM Expectation-Maximization algorithm. We set the number of Gaussian components to $K=10$ and regularize the covariances by a factor of $10^{-6}$ to prevent singularities caused by outliers. The model is trained for 100 EM iterations (or until convergence with a tolerance of $10^{-5}$). Once fit, we calculate the exact closed-form Gaussian posterior mean for each observation. 

\textit{Evaluation.} Performance is measured using the standard Mean Squared Error (MSE) between the denoised estimates $\widehat{\theta}_i$ and the clean latent truths $\theta_i$ (excluding the outliers):
$$ \text{MSE} = \frac{1}{N} \sum_{i=1}^N \|\widehat{\theta}_i - \theta_i\|^2 $$
The MSE results presented in Figure~\ref{fig:2d-exp} are representative of the average performance across 5 independent random seeds.

\subparagraph{Kernel Bandwidth Sensitivity Analysis.}\label{sec:bw_sensivity}

In this experiment, we test bandwidth sensitivity for MEC-MMD. We use data from Experiment 5 to compare performance across different bandwidth values. We use a multi-scale Gaussian kernel with bandwidths $\{{\sigma_j}\}_{j=1}^9$ selected via the median heuristic. This provides the user with a principled data-driven default. Here, we verify the sensitivity of the learned prior and downstream denoising performance on bandwidth choice. 

\begin{figure}[htbp]
    \centering
    \includegraphics[width=1\linewidth]{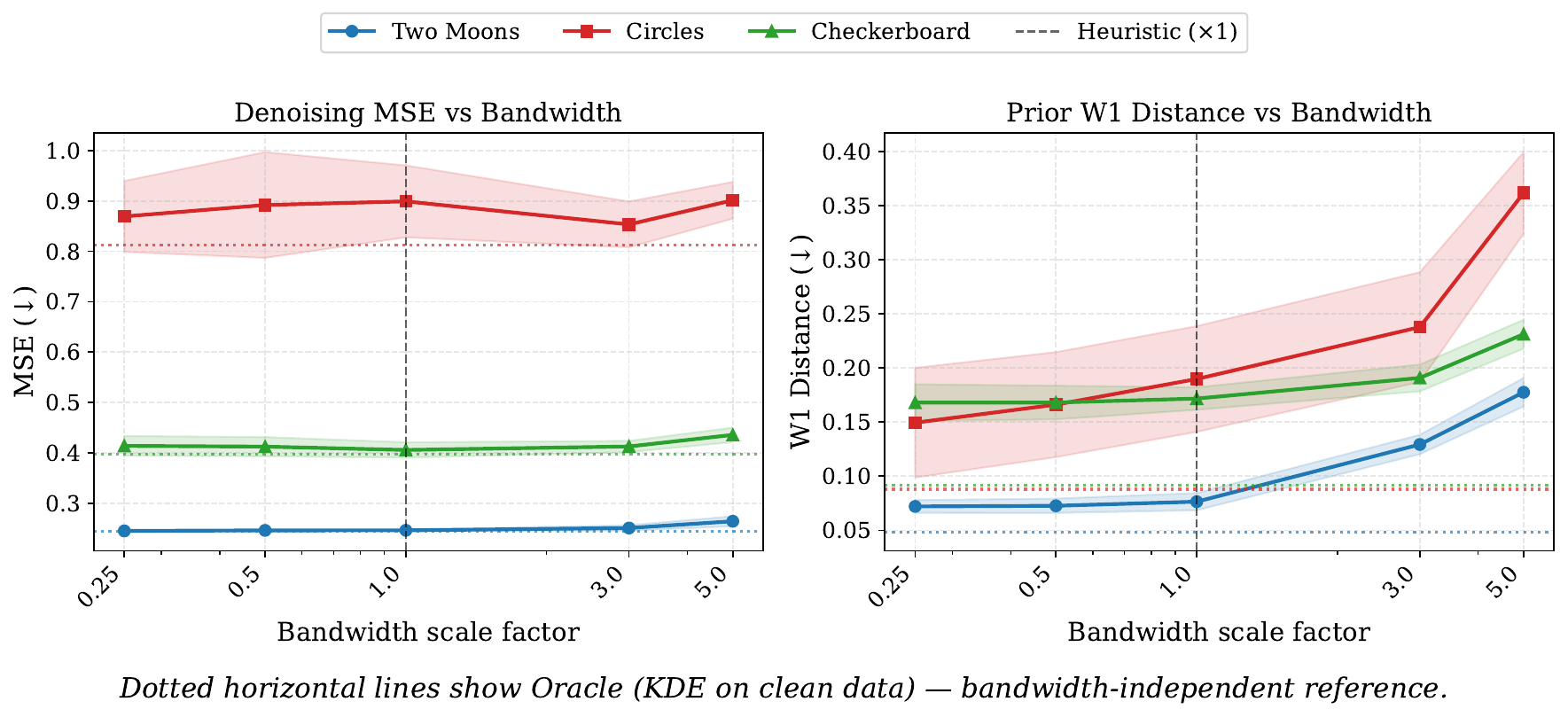}
    \caption{Kernel bandwidth sensitivity analysis.}
    \label{fig:bandwidth-sens}
\end{figure}

\textit{Setup.} We hold the full training procedure fixed and vary only a single multiplicative scale factor $s \in \{0.25, 0.50, 1.00, 3.00, 5.00\}$ applied uniformly to all nine bandwidths, i.e.\ $\tilde{\sigma}_j = s \cdot \sigma_j^{\text{heuristic}}$. The value $s = 1$ recovers the default heuristic. For each $(s, \text{dataset}, \text{seed})$ triplet, we train a fresh normalizing flow from the same initialization using the scaled bandwidths and evaluate on the held-out test set of 2000 points. Base bandwidths are computed once per seed before training the model. All other hyperparameters (architecture, learning rate, and number of gradient steps) match those of the main experiment. Results are averaged over three seeds, and we report the mean $\pm$ standard deviation.
As a bandwidth-independent reference, we include the Oracle baseline (KDE fitted on a large clean sample with no outliers), whose MSE and W1 do not depend on $s$ and are shown as horizontal dotted lines in Figure~\ref{fig:bandwidth-sens}.

\textit{Results.} Figure~\ref{fig:bandwidth-sens} plots test MSE (left) and prior W1 distance (right) as a function of $s$ for all three datasets, with shaded $\pm$1 std bands across seeds. We see that performance is remarkably stable for $s \in [0.25, 2.00]$ for MSE. MSE remains within one standard deviation of the $s = 1$ value across this range, indicating that MEC-MMD is not sensitive to moderate variations of the bandwidth around the median heuristic for denoising. However, in case of prior estimation, the performance is more sensitive to bandwidth scale. The performance remains stable for $s \in [0.25, 1.00]$ but as bandwidth scale take extreme values, the performance degrades. These results justify using the median heuristic as a fixed, tuning-free component of the MEC-MMD pipeline.

\paragraph{Experiment 6 -- Image Denoising.}
    We now evaluate the efficacy of MEC-MMD for denoising noisy images. In this experiment, we observe a dataset of corrupted images $\mathcal{D}_{obs} = \{\tilde{x}_i\}_{i=1}^N$, where $\tilde{x} = x + u$, and $u$ is drawn from a known noise distribution $p(u)$. We also have an audit sample which contains both clean and noisy images $\mathcal{D}_{\text{audit}} = \{{x}_j, \tilde{x}_j\}_{j=1}^m$. Our framework operates in two stages:
\begin{itemize}
    \item Prior Learning: We train a generative model $G_\theta(z)$ to map latent variables $z \sim \mathcal{N}(0, I)$ to the clean data manifold. The generator is optimized by minimizing the MEC-MMD between the empirical distribution of the generated noisy samples and the observed noisy dataset.
    \item Empirical Bayes Inference: Once $G_\theta$ successfully models the clean prior $q_\theta(x)$, we denoise novel test images by estimating the posterior mean $\mathbb{E}[x | \tilde{x}]$ via Importance Sampling (IS) over the learned generative manifold. 
\end{itemize}
We benchmark our methods on the MNIST dataset. We evaluate performance across two distinct noise regimes: (1) Additive White Gaussian Noise (AWGN) with $\sigma = 0.5$ and (2) Spatially Correlated Noise (generated by applying a spatial Gaussian filter to AWGN). The filtered noise is re-normalized to maintain the exact total noise variance ($\sigma=0.5$), ensuring the noise energy remains constant while the spatial structure varies from independent to highly correlated. 

\textit{Network Architecture.}
We utilize a standard Convolutional Generator. The latent vector $z \in \mathbb{R}^{128}$ is projected via a fully-connected layer to a $256 \times 7 \times 7$ feature map, followed by three transposed convolutional blocks (kernel size 4, stride 2, padding 1) with SELU activations and BatchNorm. The final layer uses a $3 \times 3$ convolution with a Sigmoid activation to output $28 \times 28$ images. 

\textit{Noise.}
We model the corruption as $y = x + n$. We utilize a 2D separable Autoregressive AR(1) Gaussian random field. The spatial correlation is controlled by $\rho \in [0, 1)$. The discrete Power Spectral Density (PSD) for an $H \times W$ grid is defined as:
$$ S(\omega_x, \omega_y) = \frac{1}{(1 - 2\rho \cos\omega_x + \rho^2)(1 - 2\rho \cos\omega_y + \rho^2)} $$
where $\omega_x = \frac{2\pi i}{H}$ and $\omega_y = \frac{2\pi j}{W}$. 
To generate a noise sample, we draw standard white noise $w \sim \mathcal{N}(0, \mathbf{I})$, transform it via 2D FFT, scale it by $\sqrt{S(\omega_x, \omega_y)}$, and apply the inverse FFT. Finally, the noise field is globally scaled such that its standard deviation exactly matches $\sigma_{\text{noise}} = 0.5$. 

\textit{Kernel Selection.}
We utilize an unbiased squared MMD formulation using a mixture of Radial Basis Function (mix-RBF) kernels:
$$ k(x, y) = \sum_{s \in \mathcal{S}} \exp \left( - \frac{||x - y||^2}{2\sigma_s^2} \right) $$
In pixel space, the bandwidths were set using the median bandwidth heuristic. 

\textit{Empirical Bayes Inference.}
Denoising is formulated as finding the posterior mean $\hat{x} = \mathbb{E}_{p(x|\tilde{x}_{obs})}[x]$. Because the true posterior is intractable, we approximate it using Importance Sampling over $K$ candidate samples generated by the trained prior $q_\theta$:
$$ \hat{x} \approx \sum_{k=1}^K w_k x_{gen}^{(k)}, \quad \text{where} \quad w_k = \frac{p(\tilde{x}_{obs} | x_{gen}^{(k)})}{\sum_{j} p(\tilde{x}_{obs} | x_{gen}^{(j)})} $$
We draw $K = 30,000$ samples per test image. The likelihood $p(\tilde{x}_{obs} | x_{gen})$ is computed assuming a known Gaussian noise model. To prevent numerical underflow in high dimensions, weights are computed in log-space using the LogSumExp trick before softmax normalization.

\textit{Training Hyperparameters.}
All models were trained using the Adam optimizer with a learning rate of $5 \times 10^{-4}$. The DCGAN was trained for 40,000 iterations with a batch size of 1,000.  All experiments were executed on an NVIDIA H200 GPU.

\paragraph{Experiment 7 -- Deconvolution across dimensions.}
\textit{Data.} We assume that the latent data lies on a one-dimensional spiral (the ``tangled ribbon''): $t\sim\mathrm{Unif}(1.5\pi,4.5\pi)$ and $(x,y)=(t\cos t,\,t\sin t)/(4.5\pi)$. The curve is embedded in $\mathbb R^d$ by appending $d-2$ zero coordinates and applying a random orthogonal matrix drawn for each dataset, with $d\in\{3,7,11,15,20,30\}$. Observations are $\widetilde Y_j=Y_j+\sigma_j\xi_j$ with $\xi_j\sim\mathcal N(0,I_d)$ and $\sigma_j\sim\mathrm{Unif}(0.0848,0.1272)$ (mean $0.106$, $\pm20\%$). We note that the noise level does not change with $d$, so the total noise power grows with $d$. Each seed draws $N=2000$ noisy training points and $10{,}000$ held-out latent points used for evaluation.

\textit{Audit and methods.} A simple random audit of $m=400$ of the $N=2000$ training points ($20\%$) is shared by all noise-aware methods and no method is given the true noise level. Na\"ive MMD ignores the measurement error and uses no audit. convMMD bootstraps the audited residual vectors $\widetilde Y_j-Y_j$. XDGMM (25 components, 100 EM iterations) uses the noise covariance $\hat\sigma^2I_d$, with $\hat\sigma$ the standard deviation of the audited residual entries. MEC-MMD applies the audit correction of Section~\ref{sec:theory} with the same audit.

\textit{Model and training.} All MMD-based methods fit the same normalising flow (4 rational-quadratic-spline blocks, 16 bins, hidden width 64, tail bound 1) with the Laplace kernel $k(x,y)=\exp(-\|x-y\|_1/h_q)$ averaged over nine bandwidths $h_q=\sqrt{Q_q}/\log_{10}(2n)$, $q=0.1,\dots,0.9$, computed once at initialisation from pairwise distances between the observed sample and model samples (augmented with bootstrapped residuals for convMMD). Training uses 3000 full-batch epochs of AdamW (learning rate $10^{-2}$ with cosine decay, weight decay $0.1$, gradient clipping at $1$).

\textit{Evaluation.} We report the sliced Wasserstein distance (Euclidean ground cost, 1000 random projections) between the held-out latent points and $10{,}000$ samples from the fitted model, multiplied by $\sqrt d$ and averaged over 5 seeds.

\section{Non-Uniform Audits}
\label{sec:nonuniform_main}
Simple random auditing (Assumption~\ref{ass:srs_audit}) can fail in cases where audits are stratified, outcome-dependent, or enriched for uncertain units. Let $\pi_j=\Pr(j\in L)$ and $\pi_{jj'}=\Pr(j,j'\in L)$ be the first- and second-order inclusion probabilities of the audit design, assumed known and positive. Replacing the factors $N/m$ and $N(N-1)/[m(m-1)]$ in \eqref{eq:Apq_me_hat} and \eqref{eq:Aqq_me_hat} by $1/\pi_j$ and $1/\pi_{jj'}$ gives an inverse-probability-weighted estimator $\widehat{\mathrm{MMD}}^2_{\mathrm{IPW}}$, which is design-unbiased for $\widehat{\mathrm{MMD}}^2_\star$ by the same Horvitz--Thompson argument as Theorem~\ref{thm:me_unbiased_sums}(Appendix~\ref{sec:nonuniform-audit}). Because small $\pi_j$ produce large weights, our experiments use the self-normalised (H\'ajek) version, which divides the audit corrections by the realised weight sums instead of $N$ and $N(N-1)$. It coincides with \eqref{eq:mmd_me} when all weights are equal and trades exact unbiasedness for lower variance. We note that the variance and CLT results above are proved for simple random auditing only.

\section{Suggestions for Future Direction and Improvements}
\label{sec:future_directions}

\paragraph{Utilizing the characteristics of ME.} The plug-in proxy $k(X_i,\widetilde Y_j)$ and $k(\widetilde Y_j,\widetilde Y_{j'})$ is only one possible choice. More generally, let
\[
\widetilde K_{ij}^{pq}
=
\psi_{pq}(X_i,\widetilde Y_j),
\qquad
\widetilde K_{jj'}^{qq}
=
\psi_{qq}(\widetilde Y_j,\widetilde Y_{j'}),
\]
be \emph{any} proxy kernel predictions, possibly built from a denoiser, a deconvolution model, or an estimate $\widehat\theta$ of the noise law.

Define audited residuals
\[
R_{ij}^{pq}
:=
k(X_i,Y_j)-\widetilde K_{ij}^{pq},
\qquad
R_{jj'}^{qq}
:=
k(Y_j,Y_{j'})-\widetilde K_{jj'}^{qq}.
\]
Then the estimators
\[
\widehat A_{pq}^{(\psi)}
=
\sum_{i=1}^n\sum_{j=1}^N \widetilde K_{ij}^{pq}
+
\frac Nm \sum_{i=1}^n\sum_{j\in L} R_{ij}^{pq},
\]
and
\[
\widehat A_{qq}^{(\psi)}
=
\sum_{j\neq j'} \widetilde K_{jj'}^{qq}
+
\frac{N(N-1)}{m(m-1)}
\sum_{\substack{j,j'\in L\\ j\neq j'}} R_{jj'}^{qq}
\]
remain design-unbiased for $A_{pq}^\star$ and $A_{qq}^\star$ by exactly the same Horvitz--Thompson proof.

Thus, the hierarchical noise model $\varepsilon\sim P(\cdot\mid \phi)$, $\phi\sim \Pi$ may be used to construct \emph{better} proxy kernel terms, but the correctness of that model is not required for the validity of the audit correction. As in prediction-powered inference, the proxy affects efficiency rather than validity.

\paragraph{Adaptive interpolation and shrinkage.}
The proposed MEC-MMD estimator corresponds to applying the full audit-based correction and is therefore design-unbiased for the oracle empirical MMD. However, as discussed in Section~\ref{sec:discussion}, this unbiasedness comes at the cost of increased variance when the audit sample is small. This suggests a natural extension based on shrinkage or interpolation between the naive and corrected estimators. In particular, one may consider the family
\[
\widehat{\mathrm{MMD}}^2_{\lambda}
=
\widehat{\mathrm{MMD}}^2_{\mathrm{naive}}
+
\lambda\Big(
\widehat{\mathrm{MMD}}^2_{\mathrm{MEC}}
-
\widehat{\mathrm{MMD}}^2_{\mathrm{naive}}
\Big),
\qquad \lambda \in [0,1],
\]
which interpolates between the low-variance but biased naive estimator ($\lambda=0$) and the unbiased but higher-variance MEC-MMD estimator ($\lambda=1$). Intermediate values of $\lambda$ introduce bias but may substantially reduce mean-squared error when the audit-based correction is noisy. 

Unlike the Horvitz--Thompson correction, the choice of $\lambda$ is not dictated by unbiasedness and must instead be selected to balance bias and variance. In principle, $\lambda$ could be estimated using sample splitting, cross-validation, or bootstrap-based risk estimation, analogously to recent work on prediction-powered inference and optimally weighted estimators \cite{angelopoulos2023ppipp, kluger2025prediction}. We emphasize that this interpolation should be viewed as an efficiency-tuning mechanism rather than a correction required for unbiasedness. A systematic study of optimal weighting strategies and their theoretical guarantees is an interesting direction for future work.

\end{document}